\documentclass[letterpaper]{article} 
\usepackage{aaai24}  
\usepackage{times}  
\usepackage{helvet}  
\usepackage{courier}  
\usepackage[hyphens]{url}  
\usepackage{graphicx} 
\usepackage{natbib}  
\usepackage{caption} 
\usepackage{epsfig}
\usepackage{graphicx}
\usepackage{amsmath}
\usepackage{amssymb}
\usepackage{multirow}
\usepackage{xcolor}
\usepackage{algorithm}
\usepackage{algorithmic}

\usepackage{newfloat}
\usepackage{listings}
\DeclareCaptionStyle{ruled}{labelfont=normalfont,labelsep=colon,strut=off} 
\floatstyle{ruled}
\newfloat{listing}{tb}{lst}{}
\floatname{listing}{Listing}
\title{Scale-Aware Crowd Count Network with Annotation Error Correction}
\author {
    Yi-Kuan Hsieh\textsuperscript{\rm 1},
    Jun-Wei Hsieh\textsuperscript{\rm 1},
    Xin li\textsuperscript{\rm 2},
    Ming-Ching Chang\textsuperscript{\rm 2},
    Yu-Chee Tseng\textsuperscript{\rm 1}
}
\affiliations {
    \textsuperscript{\rm 1}College of Artificial Intelligence and Green Energy, National Yang Ming Chiao Tung University\\
    \textsuperscript{\rm 2}Computer Science Department, University at Albany, SUNY, NY, USA.\\
}

\usepackage{bibentry}

\begin{document}
\maketitle
\begin{abstract}
Traditional crowd counting networks suffer from information loss when feature maps are downsized through pooling layers, leading to inaccuracies in counting crowds at a distance.  Existing methods often assume correct annotations during training, disregarding the impact of noisy annotations, especially in crowded scenes. Furthermore, the use of a fixed Gaussian kernel fails to account for the varying pixel distribution with respect to the camera distance. To overcome these challenges, we propose a Scale-Aware Crowd Counting Network (SACC-Net) that introduces a ``scale-aware'' architecture with error-correcting capabilities of noisy annotations. For the first time, we {\bf simultaneously} model labeling errors (mean) and scale variations (variance) by spatially-varying Gaussian distributions to produce fine-grained heat maps for crowd counting. Furthermore, the proposed adaptive Gaussian kernel variance enables the model to learn dynamically with a low-rank approximation, leading to improved convergence efficiency with comparable accuracy. The performance of SACC-Net is extensively evaluated on four public datasets: UCF-QNRF, UCF CC 50, NWPU, and ShanghaiTech A-B. Experimental results demonstrate that SACC-Net outperforms all state-of-the-art methods, validating its effectiveness in achieving superior crowd counting accuracy.

\end{abstract}

\vspace{-0.4cm}
\section{Introduction}

Crowd counting is an increasingly important technique in computer vision with applications in public safety and crowd behavior analysis \cite{li2021approaches,gao2020cnn}. Over the years, many CNN-based crowd counting methods have been developed that predict crowd density maps from a given image~\cite{li2018csrnet,xu2019autoscale,bai2020adaptive, ma2019bayesian,xiong2019open,varior2019multi,jiang2020attention,thanasutives2021encoder, zhu2019dual}. 
The total number of people in the image is then calculated by summing up the predicted values on the density map. In the past, the image was passed directly through a backbone and its last layer was used to predict the density map. However, most of existing methods did not account for the scale problem in crowd counting properly: people at the far end tend to look smaller than those at the near end. Existing counting methods have difficulty generating fine-grained density maps to accurately count people at the far end of an input image after it passes through the pooling layer.

\begin{figure}[t]
\centerline{
  {\footnotesize (a)}
  \includegraphics[width=0.23\textwidth]{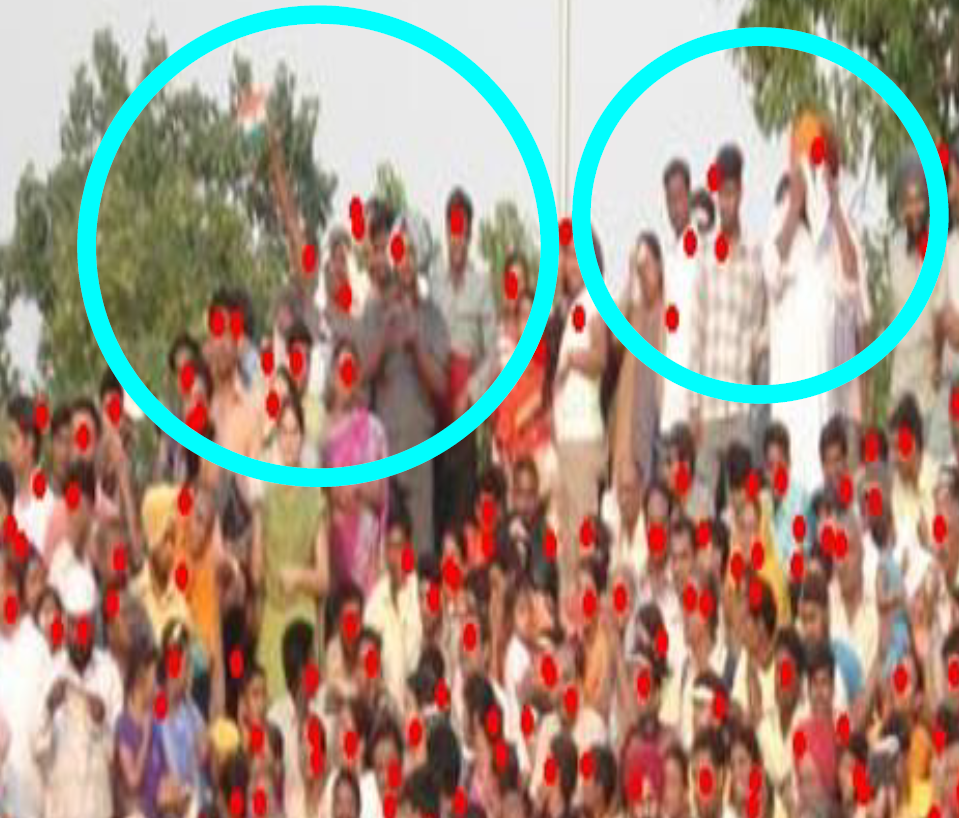}
  {\footnotesize (b)}
  \includegraphics[width=0.19\textwidth]{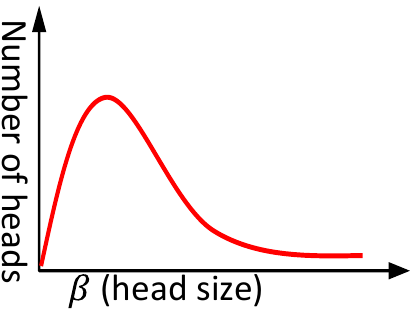}  
\vspace{-2mm}
}
\caption{Modeling uncertainty for the task of crowd counting.
(a) Inaccurate annotations lead to biased mean (red dots deviate from the center of human faces).
(b) Different camera distance leads to a distribution of head sizes (or $\beta$ characterizing the change in variance).  It is positively skewed.\vspace{-4mm}
}
\label{fig:annotation noise}
\end{figure}

\begin{figure*}[t]
\centerline{
  \includegraphics[width=1.0\linewidth]{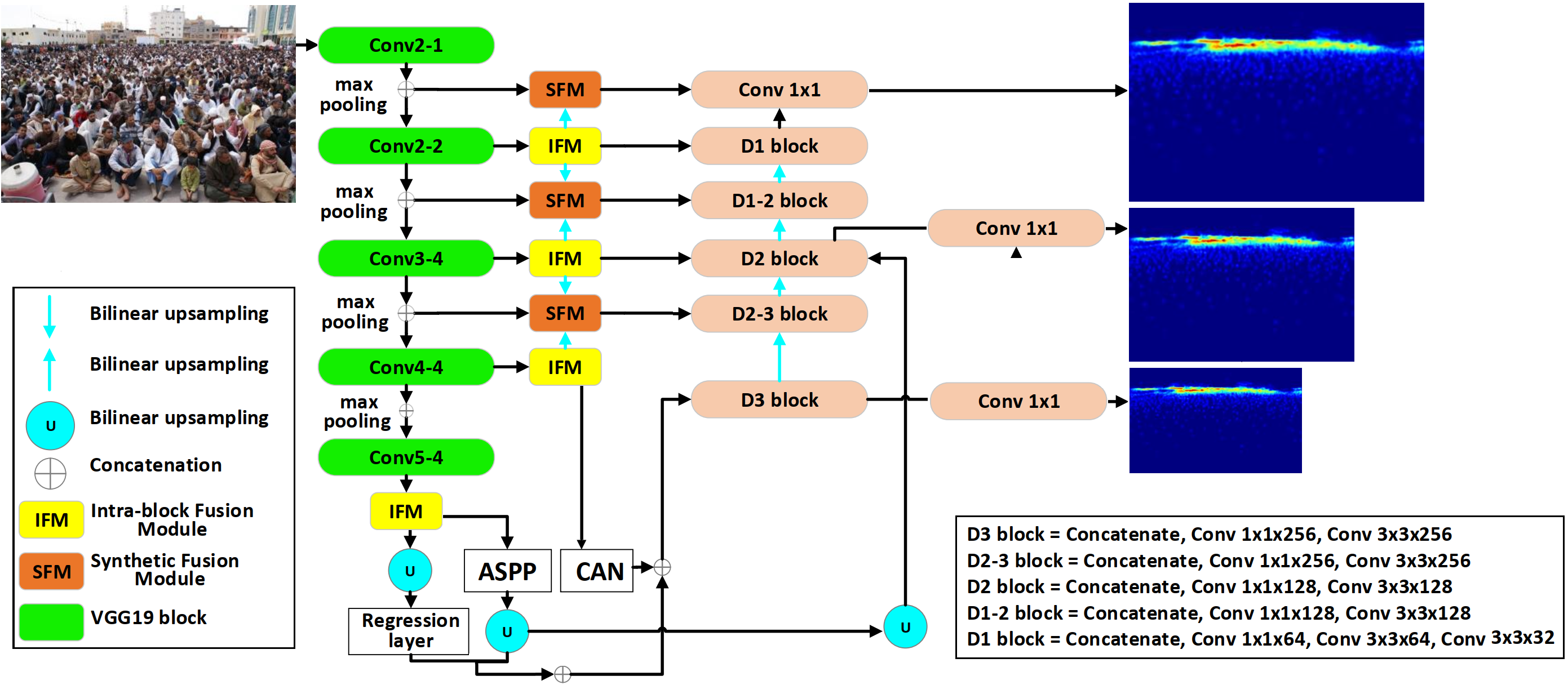}
  \vspace{-2.5mm}
}
\caption{Details of the proposed {\bf Scale-Aware Crowd Counting Network (SACC-Net)} architecture for scale-aware crown counting. With a new scale-aware loss function, it outperforms all SoTA methods on four popular crowd counting datasets.}
\label{fig:Scale-IFCnet}
  \vspace{-4mm}
\end{figure*}

Moreover, many of these methods require precise annotations from which a density map can be constructed using L2-norm~\cite{li2018csrnet,wan2019adaptive,cao2018scale} or Bayesian Loss (BL)~\cite{ma2019bayesian}. 
Unfortunately, even for human annotators, the presence of annotation errors is inevitable because ground-truth labeling might vary from subject to subject. 
As illustrated in Fig.~\ref{fig:annotation noise}, accurately pinpointing the center of each individual's head in an image is nontrivial and can pose technical challenges, particularly for people who appear small or distant. As the crowd size increases, the distance from a person to the camera is not constant: individuals situated at a far distance might only occupy a few pixels in the image, rendering annotation more challenging and unreliable. Therefore, treating all pixels equally in Bayesian loss (BL) is likely to hurt the accuracy of crowd counting. How to handle both scale variations and annotation errors in crowd counting is still an open problem, to the best of our knowledge. 

The motivation for this work is to improve the accuracy of crowd counting by addressing not only the scaling truncation problem (caused by the pooling operations), but also the problem of annotation errors across scales.  The Feature Pyramid (FP) can capture an object's visual features from coarse to fine scales and has become the standard component for most State-of-The-Art (SoTA) object counting frameworks~\cite{li2018csrnet,xu2019autoscale,bai2020adaptive, ma2019bayesian,xiong2019open,varior2019multi,jiang2020attention,thanasutives2021encoder, zhu2019dual}. However, the adopted pooling operations aim to scale the feature maps in the FP to $\frac{1}{2}$, $\frac{1}{4}$, or $\frac{1}{8}$ of the input size, where the scale truncation causes small objects to disappear dramatically. To tackle this problem, we propose a novel {\bf Synthetic Fusion Module} (SFM) to scale the feature map to $\frac{1}{2}$, $\frac{1}{3}$, $\frac{1}{4}$, $\frac{1}{6}$, {\em etc.} Then, a smoother scale space can then be obtained for fitting the ground truth whose scale changes continuously.  We propose an {\bf Intra-block Fusion Module} (IFM) to allow all feature layers within the same convolution block to be fused, so that more fine-grained information can be sent to the decoder for effective crowd counting. Finally, most existing crowd counting architectures \cite{li2018csrnet,xu2019autoscale,bai2020adaptive,xiong2019open,varior2019multi,jiang2020attention} cannot meet the operating speed requirements for real-time crowd counting.  Our architecture can be easily converted to a lightweight version with real-time efficiency and comparable accuracy.


To address the problem of annotation errors, we propose a novel {\bf scale-aware loss function} that simultaneously considers annotation noise, head-to-head correlation, and adjustment for variances at different scales.  In \cite{wan2020modeling}, a multivariate Gaussian distribution was used to solve this annotation problem.  This model is fixed for all objects of different sizes. In real images, the sizes of human heads vary at different positions.  Thus, we argue that annotation correction should be scale-aware and capable of adapting to changes in head size. To model the correlation between pixels at different scales, we derive a multivariate Gaussian distribution with a full covariance matrix of different scales. To speed up computation, we adopt a low-rank approximation method. Finally, our scale-aware loss function is designed to correct human annotation errors, so that our trained model can achieve SoTA performance in crowd counting.
Our new architecture, {\bf Scale-Aware Crowd Counting Network (SACC-Net)} as in Fig.~\ref{fig:Scale-IFCnet} is integrated into VGG-19 and trained by a new loss function with scale-aware annotation error correction that achieves SoTA performance on four popular crowd counting datasets. The main contributions of this paper are summarized as follows:

\begin{itemize} \itemsep -.1em

\item We propose SACC-Net that integrates information across layers and corrects annotation errors across scales to achieve SoTA crowd counting performance.

\item Based on an observation that the distribution of head sizes in an image is generally skewed, we create a new scale-aware density model to handle the counting annotation errors while addressing the scale variations. A new scale-aware loss function is proposed to simultaneously model scale variations and annotation errors, such that a fine-grained heat maps for crowd counting is produced.

\item An SFM is proposed to generate a smoother scale space to deal with the problem of scale truncation.  

\item An IFM is proposed to fuse all feature layers within the same convolution block to generate finer-grained information for more effective crowd counting.

\end{itemize}

\section{Related Works}


{\bf Scale variations in crowd counting:}
One critical challenge of crowd counting based on summing the density maps is the scale variation due to various distances between the cameras and the targets. To improve generalizability, \cite{zhang2015cross} proposed a CNN architecture based on a switching strategy to perform an alternative optimization between density estimation and count estimation. In the multi-column CNN \cite{zhang2016single}, each column uses a different combination of convolution kernels to extract multi-scale features. However, \cite{li2018csrnet} shows that similar features are often learned in each column of this network; therefore, the model cannot be efficiently trained as the layers become deeper. In \cite{li2018csrnet}, multi-scale features are obtained using VGG16 and convolutions are adopted with different dilation rates. Instead of using different conv kernel sizes in each layer, a multi-branch strategy is used in \cite{varior2019multi} to choose convolution filters with a fixed size while extracting multi-scale features across layers. To avoid repeatedly computing convolutional features, multi-resolution feature maps are generated by dividing a dense region into sub-regions in \cite{xiong2019open}. In \cite{liu2019context}, scale variation is handled by encoding multi-scale contextual information into the regression model. In \cite{jiang2020attention}, a density attention network generates various attention masks to focus on a particular scale. A densely connected architecture is used in \cite{miao2020shallow} to maintain multi-scale information. 


The {\bf point-wise} or {\bf dotted annotation} is widely used in most crowd counting datasets to represent each object in the image. Since no size information is included, the subsequent deviation and performance evaluation compared to the bounding-box annotation is profoundly affected. To this end, in \cite{zhang2016single}, the average distance from each head to its three neighbors is calculated, and then the head size is estimated as the Gaussian standard deviation. Synthetic crowd scenes can be generated simultaneously with annotation in \cite{wang2019crowd}. In \cite{cheng2022rethinking}, various locally connected Gaussian kernels are used to replace the original convolution filter.

{\bf Loss Function:} Traditionally, density estimation-based crowd counting approaches used the pixel-wise Mean Square Error (MSE) loss for training. More recently, alternative loss functions are developed to address the limitations of MSE loss. For example, \cite{jiang2019crowd} used a combinatorial loss including the spatial abstraction and spatial correlation terms to reduce the annotation deviation. The Bayesian loss in \cite{ma2019bayesian} leverages a density contribution probability model to mitigate the impact of deviation, though false positives still cannot be successfully reduced. The DM-count loss in \cite{wang2020distribution} measures the similarity between the predicted and ground-truth density maps. In \cite{wan2020modeling}, a multivariate Gaussian distribution-based loss function considers annotation noise and correlation, but the design is not scale aware. In practice, annotation pixel shifts or errors may not affect the counting of large objects but can significantly degrade the counting of small objects. We argue that the loss function used to correct such annotation errors should be scale aware.  

\vspace{-2mm}
\section{The Proposed Architecture and Method}

\subsection{Density Map Generation}
\label{sec:DM}

Traditional methods cast the counting task as a density regression problem~\cite{lempitsky2010learning,Sindagi_2017_ICCV, wan2020modeling}. For a given image $\mathcal{I}$ with $N$ people that we aim to count, let ${\mathrm{\textbf{H}}}_{i}$ denote the true position of the $i^{th}$ person. For any pixel location $x$ in the image $\mathcal{I}$, we model the crowd density $y$ at $x$ as a Gaussian kernel centered at each annotation point. Let $\beta$ denote the annotation variance of the Gaussian kernel and $\sum_{i=1}^{N}\mathcal{N}(x|\mu, \Sigma)$ be the Probability Density Function (PDF) for a multivariate Gaussian with mean $\mu$ and covariance matrix $\Sigma$.
We calculate the squared Mahalanobis distance as $\left \| x \right \|^{2}_{\Sigma }=X^{T}\Sigma ^{-1}X$, where $X$ is the feature vector of $x$ extracted from a network backbone.
The crowd density $y$ at position $x$ is calculated as:
\begin{equation}
    y(x) = \sum_{i=1}^{N}\mathcal{N}(x|{\mathrm{\textbf{H}}}_{i},\beta\mathrm{\textbf{I}})=\sum_{i=1}^{N}\frac{1}{\sqrt{2\pi}\beta}exp(-\frac{|| x-{\mathrm{\textbf{H}}}_{i} ||_{\beta\mathrm{\textbf{I}}}^{2}}{2}),
\label{eq:Gaussain}
\end{equation}
For all annotated head positions $\{{\mathrm{\textbf{H}}}_{i}\}^{N}_{i=1}$ in the image $\mathcal{I}$, the density map $y$ is estimated by learning a regressor $f(\mathcal{I})$ based on the L2 loss $\mathcal{L}(y,f(\mathcal{I}))$ = $\left\| y-f(\mathcal{I}) \right\|^{2}$ or a Bayesian loss~\cite{wan2020modeling}. The crowd count is calculated as the sum of the map $y$ from all pixels in $\mathcal{I}$. 

\begin{figure}[t]
\centerline{
  \includegraphics[width=0.8\linewidth]{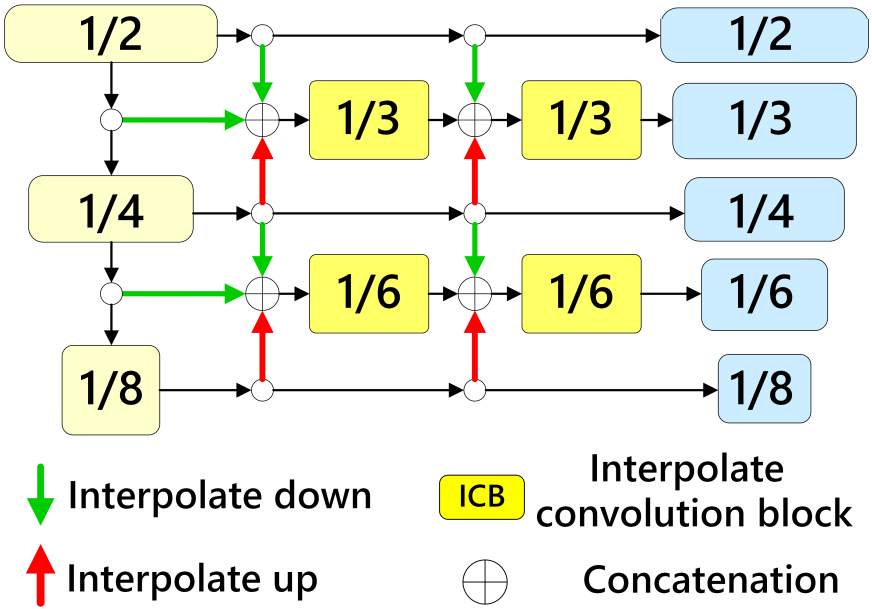}
  \vspace{-2mm}
}
\caption{The Synthetic Fusion Module (SMF) produces a smoother scaling space for heatmap generation.
}
\label{fig:SFM}
  \vspace{-1mm}
\end{figure}

\begin{figure}[t]
\centerline{
  {\footnotesize (a)}
  \includegraphics[height=0.21\textwidth]{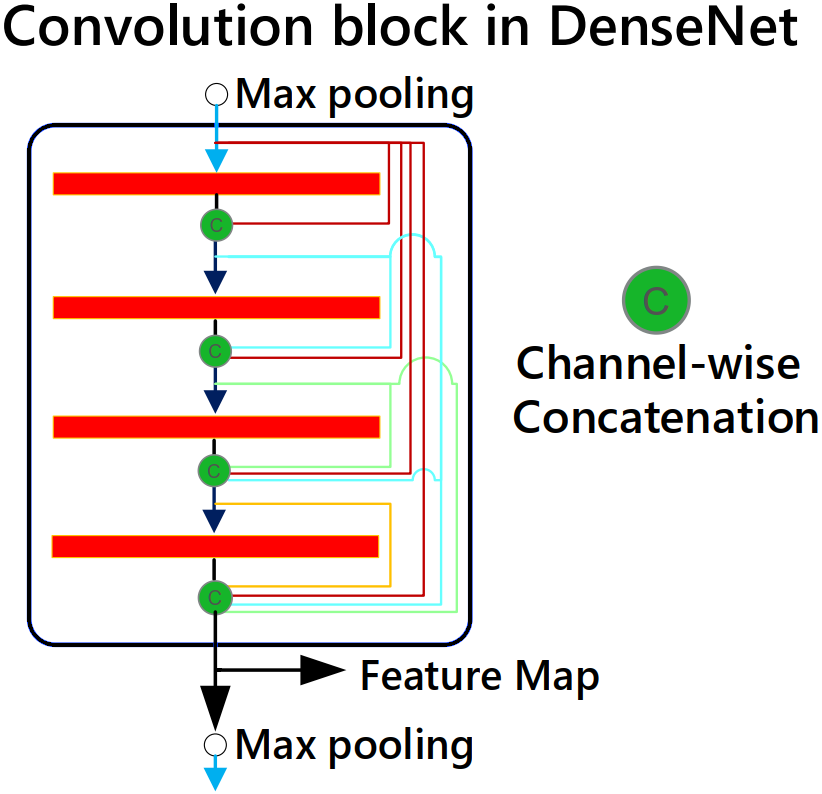}
  {\footnotesize (b)}
  \includegraphics[height=0.21\textwidth]{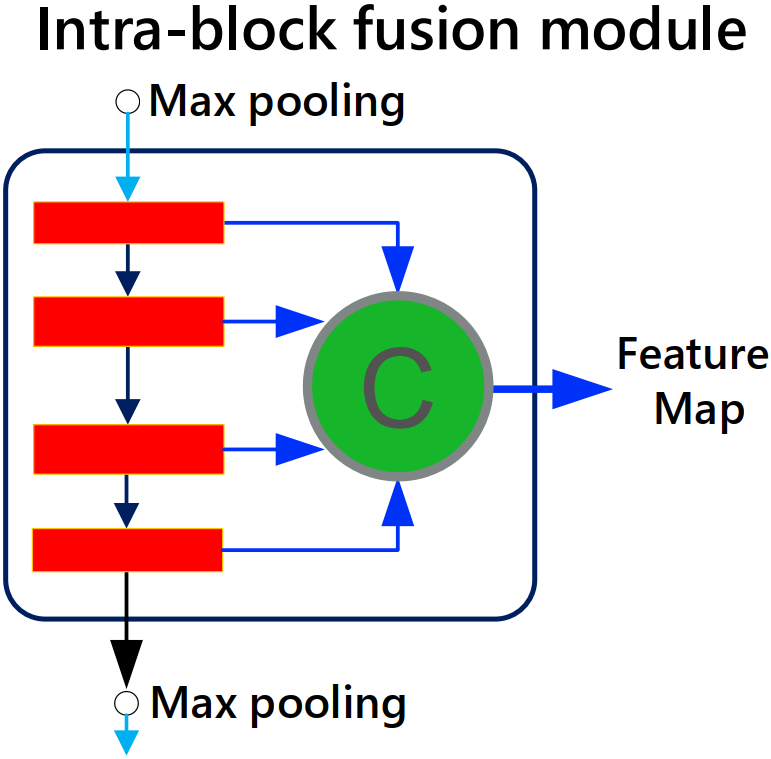}
\vspace{-2mm}
}  
\caption{
Comparison between (a) convolution block in DenseNet and (b) our Intra-block Fusion Module (IFM).
}
\label{fig:DenseNet_IFM}
\vspace{-4mm}
\end{figure}

\subsection{Scale-Aware Crowd Counting Network}
\label{sec:SACC-Net}

We next overview the Scale-Aware Crowd Counting Network (SACC-Net) architecture that generates the density map $y(x)$ as in Fig.~\ref{fig:Scale-IFCnet}. In most of CNN backbones, the  pooling or convolution with stride 2 usually down-samples the image to half and produces feature maps of $\frac{1}{2}$, $\frac{1}{4}$, $\frac{1}{8}$, and so on. We hypothesize that such {\em scale gap} is too large, which causes the features fusion within the layers to be uneven. To address this issue, we propose a novel {\em Synthetic Fusion Module (SFM)} in Fig.~\ref{fig:SFM} to generate new {\em synthetic} layers between the original layers, such that an improved set of density maps can be produced for accurate crowd counting. We further propose an {\em Intra-block Fusion Module (IFM)} in Fig.~\ref{fig:DenseNet_IFM} to allow all feature layers within the same convolution block to be fused, such that more fine-grained information can be sent to the decoder for effective crowd counting.  At the end of the SACC-Net, we adopt the ASPP~\cite{chen2017deeplab} and CAN~\cite{liu2019context} modules to leverage {\em atrous} convolutions with different rates to extract multi-scale features for accurate counting.

The {\bf Synthetic Fusion Module (SFM)} creates various synthetic layers between the original layers to scale the prediction maps to $\frac{1}{2}$, $\frac{1}{3}$, $\frac{1}{4}$, $\frac{1}{6}$ and $\frac{1}{8}$, as in Fig.~\ref{fig:SFM}. This provides a denser  scale space samples to fit the ground truth whose scale changes {\em continuously}. The synthetic layer generated by SFM can take either two or three inputs depending on its position in the SACC-Net, as shown in the brown blocks in Fig.~\ref{fig:Scale-IFCnet}. SFM first performs linearly scaling of the inputs then merge them via an $1 \times 1$ convolution. The results are then fused using a $3 \times 3$ convolution. SFM thus synthesizes a new feature layer from the two original adjacent layers, which yields a smoother scaling space for crowd counting. Details of the linear down-sampling and up-sampling operations and their time complexities are discussed in the supplementary.

{\bf Intra-block Fusion Module (IFM):}  
For most CNN architectures such as VGG, various convolutions are performed sequentially to extract features. Only the feature maps from the last layer of a convolution block are sent to the next module. In contrast, we argue that, not only the last layer but also all other layers within the convolution block can provide fine-grained features to generate an accurate density map. To reflect these ideas, our proposed IFM is different in design in comparison to the structure of the DenseNet~\cite{Huang2017DenseNet}. The convolution block of the DenseNet in Fig.~\ref{fig:DenseNet_IFM}(a) uses a fully connected structure to link all layers, which might lead to difficulties in training such as complicated computation in back-propagation, excessive RAM usage during model training, and inefficiency during inference. Fig.~\ref{fig:DenseNet_IFM}(b) shows the structure of our IFM, which uses fewer connections over DenseNet to generate the required feature maps. IFM contains three advantages over DenseNet: (1) IFM requires fewer memory usage, as it uses $1 \times 1$ convolution to directly obtain the output. (2) IFM can obtain more representative features by aggregating all layers of information. (3) IFM contains fewer parameters and is more efficient over DenseNet. Time complexity of IFM is analyzed in the supplementary. 

{\bf Light-Weight Architecture}: Density-based crowd counting methods can achieve good counting accuracy, but existing methods are inefficient for running in real-time applications. The processing of feature maps in SACC-Net goes in two routes, as shown in the \texttt{Conv-2-1} block in Fig.~\ref{fig:Scale-IFCnet}: one branch goes through a VGG block, and the other goes directly to the convolution block with two simple convolutions and move on (see Fig. 2 in the Supplementary). This separation can balance the computation load of each layer and reduce the memory load. In each convolution block, $e.g.$, the green block ``\texttt{Conv-n-x}'' in Fig.~\ref{fig:Scale-IFCnet}), only half the number of channels instead of the original convolution on all channels are sent to the next convolution block, $e.g.$, the block ``\texttt{Conv-(n+1)-x}'' in Fig.~\ref{fig:Scale-IFCnet}. This design can greatly reduce the number of model parameters while maintaining stable accuracy performance. Additional details regarding this lightweight design are provided in the supplementary. 

\subsection{Scale-Aware Annotation Noise}

We aim to deal with uncertainty in the manually labeled annotations as shown in Fig.~\ref{fig:annotation noise}(a). Our assumption is that the point-wise head annotations can inevitably come with annotation errors. This annotation errors will cause the crowd density $y$ in Eq.~\eqref{eq:Gaussain} to be incorrectly estimated. We next derive a solution to address this.

Let $\tilde{\mathrm{\textbf{H}}}_{i}$ denote the annotated head position of the $i^{th}$ person with potential annotation error, and $\varepsilon_{i}$ denote its annotation noise, $ \tilde{\mathrm{\textbf{H}}}_{i}=\mathrm{\textbf{H}}_{i}+\varepsilon_{i}$. We assume the annotation noise is independent and identically distributed (i.i.d), $\varepsilon_{i}\overset{i.i.d}{\sim }\mathcal{N}(0,\alpha\mathit{\textbf{I}})$, where $\alpha$ is a annotation variance parameter. Considering the case of annotation noise, we model the density $\mathbb{D}(x)$ at location $x$ as: 
\begin{equation}
\begin{aligned}
    \mathbb{D}(x) &=  \sum_{i=1}^{N}\mathcal{N}(x|\mathrm{\tilde{\mathrm{\textbf{H}}}_{i}},\beta\mathit{\textbf{I}})=\sum_{i=1}^{N}\mathcal{N}(x|{\mathrm{\textbf{H}}}_{i}+\varepsilon_{i},\beta\mathit{\textbf{I}})\\
    &=\sum_{i=1}^{N}\mathcal{N}(q_{i}|\varepsilon_{i},\beta\textbf{I})\cong \sum_{i=1}^{N}\phi_{i}.
    \end{aligned}
    \label{eq:density}
\end{equation}
where $\beta$ is defined
in Eq.~\eqref{eq:Gaussain}, $q_{i}=x-{\mathrm{\textbf{H}}}_{i}$ denoting the position difference between the $i^{th}$ annotation and $x$, and $\phi_{i}$ denotes the individual Gaussian kernel for the $i^{th}$ annotation. In the literature, \cite{wan2020modeling} did not distinguish the range of annotation errors among small and large objects. They proposed a fixed-scale model using the NoiseCC loss to rectify the annotation noise.
One key problem for all SoTA methods~\cite{lempitsky2010learning,Sindagi_2017_ICCV, wan2020modeling} regarding Eq.~\eqref{eq:density} is that a fixed $\beta$ with constant value is used to model $\mathbb{D}(x)$ of the crowd density around each head position. 
As aforementioned, the head sizes appear in different sizes according to their distances {\em w.r.t.} the observing camera. To this end, our design makes $\beta$ scale-aware and adaptive to the size of each head appearing in the image.

\subsection{Adaptive Gaussian Kernel}

As shown in Fig.~\ref{fig:annotation noise}, the distribution of $\beta$ is positively skewed (small heads occupy more). With this observation, this section will propose a scale-adaptive Gussian model for heat map generation and annotation error correction. Assume that there are $S$ scales used to model a head with annotation errors.  Then, Eq.~\eqref{eq:density} can be rewritten as:\vspace{-2mm}
\begin{equation}
\begin{aligned}
    \mathbb{D}(x) &=\sum_{i=1}^{N} \sum_{s=1}^{S} w_s \mathcal{N}(q_{i}|\varepsilon_{i},\beta_s\textbf{I})\cong \sum_{s=1}^{S} w_s \sum_{i=1}^{N} \phi_{i}^s, \vspace{-3mm}
    \end{aligned}
    \label{eq:all-density}
\end{equation}
where the density of a head is modeled with a mixed Gaussian model, and $\sum_{s=1}^{S} w_s=1$. In addition, $\phi_i^s=\mathcal{N}(q_{i}|\varepsilon_{i},\beta_{s}\textbf{I}) $, $i.e$., the Gaussian kernel placed in the $i$th annotation at the scale $s$ and parameterized with the annotation error $\varepsilon_{i}$ and the variance $\beta_s$.   Let $\mathbb{D}_{s}= w_s \sum_{i=1}^{N}\phi_{i}^s$. In addition, we denote $\mathcal{I}_s$ as the scaled-down version of $\mathcal{I}$ at scale $s$. For all pixels $x_j$ in $\mathcal{I}_s$, a multivariate random variable for the density map $\mathbb{D}(x)$ at the scale $s$ can be constructed as:\vspace{-2mm}
\begin{equation}
\begin{aligned}
    \mathbb{D}_{s}=[\mathbb{D}_{s}(x_1), \cdots, \mathbb{D}_{s}(x_j), \cdots , \mathbb{D}_{s}(x_{J_s})],\vspace{-2mm}
    \end{aligned}
    \label{eq:DS}
\end{equation}
where $J_s$ is the number of pixels in $\mathcal{I}_s$. 
 
\subsubsection{Scale-aware probability distribution}

To calculate 
$\mathbb{D}_{s}$ in closed form, we approximate it by a Gaussian, $i.e$., $\hat{p}(\mathbb{D}_{s})\sim\mathcal{N}(\mathbb{D}_{s}|\mu_{s}, \sigma^{2}_{s})$ with the scale-aware mean $\mu_{s}$ and variance $\sigma^{2}_{s}$.  The mean $\mu_{s}$ is calculated as: \vspace{-1mm}
\begin{equation}
\begin{aligned}
\mu_{s}&= \mathbb{E}[\mathbb{D}_{s}]=\mathbb{E}[  w_s \sum_{i=1}^{N} \mathcal{N}(q_{i}|\varepsilon_{i},\beta_{s}\textbf{I})] \\ 
&= w_s \sum_{i=1}^{N}\mathcal{N}(q_{i}|0,(\alpha+\beta_{s})\textbf{I}) = \sum_{i=1}^{N}\mu_i^s,
\label{eq:mu}
\end{aligned}
\end{equation}
where $\mu_i^s = w_s \mathcal{N}(q_{i}|0,(\alpha+\beta_{s})\textbf{I})$  and the annotation error $\varepsilon_{i}\sim\mathcal{N}(0|0, \alpha \textbf{I})$. The variance $\Sigma^{2}_{s}$ is given by:
\begin{equation}
\begin{aligned}
    \Sigma^{2}_{s}&=\mathrm{var}(\mathbb{D}_{s})=\mathbb{E}[\mathbb{D}_{s}^{2}]-\mathbb{E}[\mathbb{D}_{s}]^{2}\\ &\cong \sum_{i=1}^{N}[\frac{w_s^2}{4\pi\beta_{s}}\mathcal{N}(q_{i}|0,(\beta_{s}/2+\alpha)\textbf{I})-(\mu_i^s)^{2}].
    \label{eq:var}
\end{aligned}
\end{equation}

\subsubsection{Gaussian approximation to scale-aware joint likelihood $\textbf{D}_{s}$}

Next, the covariance term  $\mathrm{Cov}(\mathbb{D}_{s}{(x_j)},\mathbb{D}_{s}{(x_k)})$  between locations $x_j$ and $x_k$ needs to be calculated.  We model it by a multivariate Gaussian approximation of the joint likelihood $\textbf{D}_{s}$ at the scale $s$. Let $q_{i}{(x_j)}=x_j-\tilde{\mathrm{\textbf{H}}}_{i}$ be the difference between the spatial location of the $i$-th annotation and the location of the pixel $x_j$. Based on Eq.~\eqref{eq:all-density}, the density value $\mathbb{D}_{s}{(x_j)}$ is calculated as:
\begin{equation}
    \mathbb{D}_{s}{(x_j)} = w_s \sum_{i=1}^{N}\mathcal{N}(q_{i}{(x_j)}|\varepsilon_{i},\beta_s\textbf{I}) = w_s \sum_{i=1}^{N}\phi_i^s{(x_j)},
    \label{eq:Phi}
\end{equation}
where $\phi_i^s{(x_j)}=\mathcal{N}(q_{i}{(x_j)}|\varepsilon_{i},\beta_s\textbf{I})$ and annotation noise $\varepsilon_{i}$ is the same random variable across all $\phi_i^s{(x_j)}$.  Define the Gaussian approximation to $\textbf{D}_{s}$ as 
$
    \hat{p}(\textbf{D}_{s}) = \mathcal{N}(\textbf{D}_{s}|\mu_{s},\Sigma_{s}),
    \label{eq:PDS}
$
where $\mu_{s}$ and $\Sigma_{s}$ are defined in Eqs.~\eqref{eq:mu} and \eqref{eq:var}. The $j^{th}$ entry in $\mu_{s}$ is $\mathbb{E}[\mathbb{D}_{s}(x_j)]$ = $\sum_{i=1}^{N}\mu_i^s{(x_j)}$ from Eq.~\eqref{eq:mu}. The diagonal of the scale-aware covariance matrix is calculated as $\boldsymbol{\Sigma}_{x_j,x_j}^s$ = $\mathrm{Var}(\mathbb{D}_{s}{(x_j)})$. The covariance term is then: \vspace{-0.5mm}
\begin{equation}
\begin{aligned}
    \boldsymbol{\Sigma}_{x_j,x_k}^s &= \mathrm{Cov}(\mathbb{D}_{s}{(x_j)},\mathbb{D}_{s}{(x_k)})\\
    &=  \sum_{i=1}^{N}[w_s^2\Omega_i^s{(x_j,x_k)}-\mu_i^s{(x_j)}\mu_i^s{(x_k)}].
    \end{aligned}
    \label{eq:cov}
\end{equation}
where $\Omega_i^s{(x_j,x_k)}$ = $\mathbb{E}[\phi_i^s{(x_j)}\phi_i^s{(x_k})]$.

\subsection{Low-rank Approximation using SVD}

Since the dimension of $\boldsymbol{\Sigma}_{x_j,x_k}^s$ is huge, $i.e$, $J_s \times J_s$, this section will derive its low-rank approximation with its non-zero rows and columns for efficiency improvement. Let $\hat{\boldsymbol{\Sigma} }^{s}$ denote the approximation to $\boldsymbol{\Sigma}^{s}$ using Singular Value Decomposition (SVD) calculated as:

\begin{equation}
  \hat{\boldsymbol{\Sigma} }^{s} \cong \boldsymbol{\Sigma}^{s}  = \textbf{U}^{s}\textbf{C}_{L}^{s}\textbf{V}^{T},
    \label{eq:Sigma}
\end{equation}
where $\textbf{U}^{s}$ is an $J_s\times J_s$ orthogonal matrix, $\textbf{C}_{L}^{s}$ is a nonnegative $J_s\times J_s$  diagonal matrix with diagonal entries sorted from high to low, and $\textbf{V}^{T}$ is a $J_s\times J_s$ orthogonal matrix.  Let $v_j^s$ = $\boldsymbol{\Sigma}_{x_j,x_j}^s$.  To obtain this low-rank approximation, each pixel $x_j$ is first ordered by $v_j^s$. Then, the top-$M$ pixels whose percentages of variance are larger than 0.8, $i.e$. 
\begin{equation}
  \frac{\sum_{j=1}^{M}v_j^s}{\sum_{i=1}^{J} v_j^s} > 0.8,
\end{equation}
are selected from $\mathcal{I}_s$ for this low-rank approximation. Let the set of indices of the top $M$ pixels be denoted by $L$, $i.e.$, $L=\{l_1,l_2,\dots,l_m,\dots,l_{M}\}$. Then, only the elements in $L$ are selected to approximate $\boldsymbol{\Sigma}^{s}$. The approximation of a matrix $\boldsymbol{\Sigma}^{s}$ by a rank-$M$ matrix requires a representation of $\boldsymbol{\Sigma}^{s}$ as the sum of several ingredients ordered by their importance. SVD lends itself to this task by transforming $\boldsymbol{\Sigma}^{s}$ into the sum of rank-1 matrices (weighted by the corresponding singular values), namely, $\boldsymbol{\Sigma}^{s} =\textbf{U}^{s}\textbf{C}_{L}^{s}{\textbf{V}^{s}}^{T}$ is equivalent to:\vspace{-2mm}
\begin{equation}
    \boldsymbol{\Sigma}^{s} = \sum^{J_s}_{i=1}c_{i}^{s}\cdot\textbf{u}^{s}_{i}{\textbf{v}_{i}^{s}}^{T},\vspace{-2mm}
    \label{eq:SVD}
\end{equation}
where the scale $s=1,...,S$, $c_{i}^{s}$ is the $i$th singular value and $\textbf{u}^{s}_{i},\textbf{v}_{i}^{T}$ are the corresponding left and right singular vectors. A natural idea is to keep only the top $M$ terms on the right-hand side of Eq. (\ref{eq:SVD}). That is, for $\boldsymbol{\Sigma}^{s}$ as in Eq. (\ref{eq:SVD}) and a target rank $M$, the proposed rank-$M$ approximation is:\vspace{-2mm}
\begin{equation}
\hat{\boldsymbol{\Sigma} }^{s}  \cong \sum^{M}_{i=1}c_{i}^{s}\cdot\textbf{u}^{s}_{i} {\textbf{v}_{i}^{s}}^{T},\vspace{-2mm}
\label{eq:low-rank SVD}
\end{equation}
where the singular values ( $s_1 \ge s_2 \ge \cdots \ge s_{J_s} \ge 0 $) have been sorted, and $\textbf{u}^{s}_{i},\textbf{v}_{i}^{T}$ denote the $i$th left and right singular vectors. With $\hat{\boldsymbol{\Sigma} }^{s} $, the rank-M approximate negative log-likelihood function is:
\begin{equation}
 -\mathrm{log} \hat{p}(\textbf{D}_{s}) = -\mathrm{log} \mathcal{N}(\textbf{D}_{s}|\mu_{s},\hat{\boldsymbol{\Sigma}}^{s})\propto ||\textbf{D}_{s}-\mu_{s}||^{2}_{\hat{\boldsymbol{\Sigma}}^{s}}.
    \label{eq:apro_likelhood}
\end{equation}
The time complexities to calculate  the right-hand sides of Eq.(\ref{eq:low-rank SVD}) and Eq.(\ref{eq:apro_likelhood}) take $O(M^3)$ and $O(M^2)$, in contrast
to $O(J_s^3)$ and $O(J_s^2)$ required to calculate the original matrix $\boldsymbol{\Sigma}^{s}$ and the distance $||\textbf{D}_{s}-\mu_{s}||^{2}_{{\boldsymbol{\Sigma}}^{s}}$, respectively.

\subsection{Regularization and the Final Loss Term}

The Gaussian approximation to $\textbf{D}_{s}$ can be obtained
based on Eqs.~\eqref{eq:low-rank SVD} and \eqref{eq:apro_likelhood}. To ensure that the predicted density map near each annotation satisfies the density sum to 1, for the $i$-th annotation point, we define the regularizer $\mathcal{R}_{i}^{s}$ as: 
\begin{equation}
    \mathcal{R}_{i}^{s} = |\sum_{j}\mathbb{D}_{s}(x_j)\frac{\phi_{i}^s(x_j)}{\sum_{i=1}^{N} \phi_{i}^s{(x_j)}}-1|,
\end{equation}
where $\mathbb{D}_{s}(x_j)$ is the $j^{th}$ term of $\textbf{D}_{s}$.  Let  $\bar{\textbf{D}}_{s}=\textbf{D}_s-\mu_{s}$. Then, the final loss function is:\vspace{-2mm}
\begin{equation}
    \mathcal{L} =  \sum_{s=1}^{S} \bar{\textbf{D}}_{s}^{T}(\hat{\Sigma}^{s})^{-1}\bar{\textbf{D}}_{s}+\sum_{s=1}^{S}\sum_{i=1}^{N} \mathcal{R}_{i}^{s}.
     \label{eq:loss_function}
\end{equation}

\section{Experimental Results}  

We evaluated our crowd counting method and compared it with 14 SoTA methods in four public datasets, UCF-QNRF~\cite{idrees2018composition}, UCF CC 50~\cite{idrees2013multi}, NWPU-Crowd~\cite{Wang2020NWPU}, and ShanghaiTech Parts A and B~\cite{zhang2016single}.



\subsection{Model Training Parameters}

Our method was pre-trained on ImageNet~\cite{imagenet} with the Adam optimizer.  Since the image dimensions in the used datasets are different, patches with a fixed size are cropped at random locations, then randomly flipped (horizontally) with probability 0.5 for data augmentation.  The learning rates used in the training process are $1e^{-5}$, $1e^{-5}$, $1e^{-5}$, and $1e^{-4}$ for the UCF-QNRF, UCF CC 50, NWPU, and ShanghaiTech datasets, respectively.  To stabilize the training loss change, we use batch sizes 10, 10, 15, and 10, respectively. All parameters used in the training stage are listed in Table~\ref{tab:Training_p}. Similarly to other SoTA methods~\cite{li2018csrnet,xu2019autoscale,bai2020adaptive,xiong2019open,varior2019multi,jiang2020attention,thanasutives2021encoder, zhu2019dual}, the mean average error (MAE) and the mean squared error (MSE) are used to evaluate the performance of our architecture.

\subsection{Parameter Settings for $\beta_s$ and $w_s$} 
\begin{figure}[t]
\centerline{
  {\footnotesize (a)}
  \includegraphics[height=0.32\linewidth]{Figures/image_map.png}
  {\footnotesize (b)}
  \includegraphics[height=0.32\linewidth]{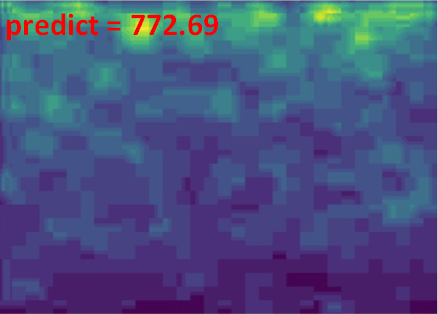}
  }
\centerline{
  {\footnotesize (c)} 
  \includegraphics[height=0.32\linewidth]{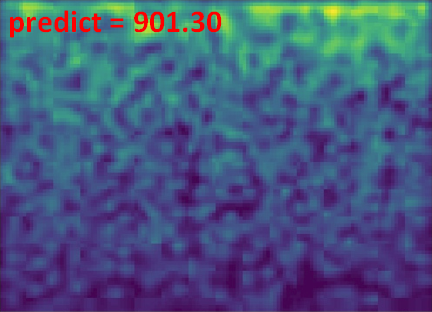}
  {\footnotesize (d)}
  \includegraphics[height=0.32\linewidth]{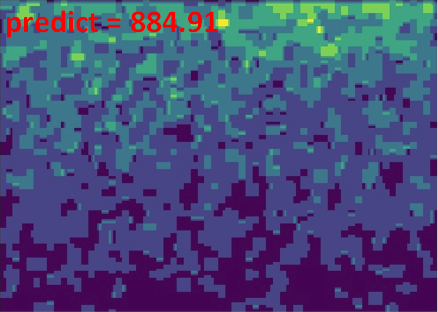}
\vspace{-2mm}
} 
\caption{Visualization crowd counting heatmaps generating using
different loss functions. (a) Input image (GT = 855, ShanghaiTech
Part A). (b) MSE loss. (c) Bayesian loss. (d) Our scale-aware loss. Apparently, our scale-aware loss generates sharper and semantically more meaningful heatmap than others (objectively, it is also the closest to the GT).
}
\label{fig:visualization_compar_loss}
\vspace{-4mm}
\end{figure}

As shown in Fig.~\ref{fig:annotation noise}(a), Fig.~\ref{fig:visualization_compar_loss}(a), and Fig.~\ref{fig:intra}(a), the number of heads of the crowd decreases according to the head size $h$. This $P_{head}(h)$ distribution is positive-skewed and can be easily obtained by accumulating from the training data. In Eq.~\eqref{eq:Gaussain}, the variance parameter $\beta$ is proportional to the head size. We can set the mean of $h$ as the initial value of $\beta_{1}$, $\beta_{1}= \sum_{h} hP_{head}(h)$.  In a CNN backbone such as VGG19, the pooling operation will reduce the feature map size by half, thus also reduce the head size in the feature map. Given $\beta_s$, the value of $\beta_{s+1}$ can be obtained in a recursive form: $\beta_{s+1}=\beta_s/2$. Our analysis yields $\beta$ to be approximately 8.3.
This subsampling operation will also
make small heads to eventually disappear.  Then, $w_s$ is set to $P_{head}(\beta_{(S+1-s)})$, where $S$ is the largest scale used to model
 $\mathbb{D}(x)$ in Eq.~\eqref{eq:all-density}. We set $S=3$. After normalization, we have $\sum_{s=1}^Sw_s=1$. 

\begin{table}[t]
\caption{Detailed parameters used for training.\vspace{-2mm}
}
\begin{tabular}{c|c|c|c}
\hline
Dataset      & learning rate & batch size & crop size       \\ \hline
UCF-QRNF     & 1e-5     & 12       & 512×512 \\
UCF CC 50    & 1e-5     & 10       & 512×512 \\
NWPU         & 1e-5     & 8       & 512×512 \\
ShanghaiTech & 1e-4     & 12       & 512×512 \\ \hline
\end{tabular}
\label{tab:Training_p}
\vspace{-1mm}
\end{table}



\subsection{Performance Comparisons {\em w.r.t.} Loss Functions} 

\begin{table}[t]
\caption{Accuracy comparisons among different loss functions
with various backbones on UCF-QNRF. \vspace{-2mm}
}
\label{tab:COM_diff loss}
\centerline{
\setlength{\tabcolsep}{0.8mm}{
\begin{tabular}{c|cc|cc|cc}
\hline
\multirow{2}{*}{} & \multicolumn{2}{c|}{VGG19} & \multicolumn{2}{c|}{CSRNet} & \multicolumn{2}{c}{MCNN} \\
                  & MAE         & MSE          & MAE          & MSE          & MAE         & MSE        \\ \hline
L2                & 98.7        & 176.1        & 110.6        & 190.1        & 186.4       & 283.6      \\
BL                & 88.8        & 154.8        & 107.5        & 184.3        & 190.6       & 272.3      \\
NoiseCC          & 85.8        & 150.6        & 96.5         & 163.3        & 177.4       & 259.0      \\
DM-count          & 85.6        & 148.3        & 103.6        & 180.6        & 176.1       & 263.3      \\
Gen-loss & 84.3        & 147.5        & 92.0         & 165.7        & 142.8       & 227.9      \\ \hline
Ours              & \textbf{83.47}       & \textbf{140.34}       & \textbf{90.83}  & \textbf{150.67}       & \textbf{134.52}      & \textbf{213.71} \\   \hline
\end{tabular}}
}
\vspace{-2mm}
\end{table}

We compare the proposed loss function against L2, BL~\cite{ma2019bayesian}, NoiseCC~\cite{wan2020modeling}, DM-count~\cite{wang2020distribution}, and the generalized loss~\cite{Wan_2021_CVPR} under different backbones to evaluate the effectiveness. Table~\ref{tab:COM_diff loss} shows the performance evaluation results. Clearly, our proposed scale-aware loss function outperforms other SoTA loss functions under various backbones. Since human head sizes are different, the same annotation error causes different effects to affect the accuracy of crowd counting. Although NoiseCC~\cite{wan2020modeling} has pointed out that the annotation noise will affect the accuracy of crowd counting, their work did not address the scaling issues.  Our scale-aware loss function can properly handle it and outperforms other losses on UCF-QNRF.

\subsection{Comparisons with SoTA Methods} 

\begin{table*}[t]
\caption{Performance comparisons among different SoTA crowd counting methods.
\vspace{-2mm}
}
\label{tab:COM_ALL}
\centerline{
\setlength{\tabcolsep}{0.2mm}
\begin{tabular}{c|c|cc|cc|cc|cc|cc}
\hline
\multirow{2}{*}{Methods}  &\multirow{2}{*}{Venue} &\multicolumn{2}{c|}{UCF-QNRF} & \multicolumn{2}{c|}{NWPU} & \multicolumn{2}{c|}{S. H. Tech-A} & \multicolumn{2}{c|}{S. H. Tech-B} & \multicolumn{2}{c}{UCF CC 50} \\
                             &      & MAE           & MSE           & MAE            & MSE           & MAE                & MSE                 & MAE                & MSE         & MAE                & MSE         \\ \hline
CSRNet      &CVPR'18                       & -             & -             & 121.3          & 522.7         & 68.2               & 115.0               & 10.3               & 16.0         & 266.1             & 397.5        \\
CAN &CVPR’19  & 107  & 183  & -  & - & 62.3               & 100.0               & 7.8                & 12.2         & 212.2             & 243.7        \\
S-DCNet    &ICCV’19                        & 104.4         & 176.1         & -      & -     & 58.3               & 95.0                & 6.7                & 10.7        & 204.2             & 301.3         \\
SANet&ECCV’18   & -  & - & 190.6  & 491.4 & 67.0  & 104.5  & 8.4                & 13.6          & 258.4             & 334.9       \\
BL  &ICCV’19                              & 88.7          & 154.8         & 105.4          & 454.2         & 62.8               & 101.8               & 7.7                & 12.7         & 229.3             & 308.2        \\
SFANet&-  & 100.8          & 174.5         & -          & -         & 59.8               & 99.3                & 6.9                & 10.9       & -             & -          \\
DM-Count&NeurIPS’20                      & 85.6          & 148.3         & 88.4          & 498.0         & 59.7               & 95.7                & 7.4                & 11.8        & 211.0             &  291.5         \\
RPnet &CVPR’15                          & -          & -         & -     & -         & 61.2   & 96.9   & 8.1   & 11.6       & -             & -          \\
AMSNet&ECCV’20  & 101.8  & 163.2         & -     & -         & 56.7.2   & 93.4   & 6.7   & 10.2     & 208.4             &  297.3            \\
M-SFANet&ICPR'21                          & 85.6          & 151.23        & -         & -             & 59.69              & 95.66              & 6.38               & 10.22         & 162.33             & 276.76       \\
TEDnet&CVPR’19                             & 113.0         & 188.0         & -          & -         & 64.2               & 109.1               & 8.2                & 12.8        & 249.4             & 354.5         \\
P2PNet &ICCV’21                            & 85.32         & 154.5         & 77.44          & 362         & 52.74               & 85.06               & 6.25                & 9.9         & 172.72             & 256.18        \\
	
GauNet&CVPR’22                            & 81.6         & 153.7          & -          & -         & 54.8               & 89.1               & 6.2                & 9.9          &  186.3             & 256.5       \\
MAN &CVPR’22                            & 77.3/83.4$^*$         & 131.5/146$^*$          & 76.5/76.6$^*$          & 323.0/465.4$^*$         & 56.8               & 90.3               & -                & -          &  -             & -    \\  

\hline

SACC-Net(BL Loss) &-                          & 85.42         & 145.44        & 86.72         & 442.9        & 55.28              & 90.37               & 6.5               & 10.68         & 167.48             & 235.41       \\ 
SACC-Net(our loss) &-                             &\textbf{77.12}         & \textbf{124.25}         & \textbf{75.52}          & \textbf{349.73}         & \textbf{52.19}               & \textbf{76.63}               & \textbf{6.16}                  & \textbf{9.71}  & \textbf{150.66}             & \textbf{187.89}\\ 
\hline
\end{tabular}
}
\centerline{
Symbol $^*$ denotes scores produced by running the original source codes provided by the authors.
}
\vspace{-4mm}
\end{table*}

To further evaluate the performance of our proposed method, 14 SoTA methods are compared here for performance evaluation; that is,
CSRNet~\cite{li2018csrnet}, CAN~\cite{liu2019context}, S-DCNet~\cite{xiong2019open}, SANet~\cite{cao2018scale}, BL~\cite{ma2019bayesian}, SFANet~\cite{zhu2019dual}, DM-Count~\cite{wang2020distribution}, RPnet~\cite{zhang2015cross}, AMSNet~\cite{Hu2020NAS} M-SFANet~\cite{thanasutives2021encoder}, TEDnet~\cite{jiang2019crowd}, P2PNet~\cite{song2021rethinking}, GauNet~\cite{cheng2022rethinking}, and MAN~\cite{Lin_2022_CVPR}. Table~\ref{tab:COM_ALL} shows 
the comparative results among these SoTA methods on four benchmark datasets. Clearly, our method achieves the best MAE on all the above datasets, especially for large-scale datasets such as UCF-QNRF, NWPU-Crowd, and ShanghaiTech Part A. As to the MSE metric, our method outperforms all SoTA methods except MAN~\cite{Lin_2022_CVPR}. 

\begin{table}[t]
\caption{Ablation study of SACC-Net for running SFM+IFM at different density scales on UCF-QNRF.
\vspace{-2mm}
}
\label{tab:ablation}
\centerline{
\setlength{\tabcolsep}{1.9mm}
\begin{tabular}{c|ccc|cc}
\hline
 \multirow{2}{*}{SFM+IFM} & \multirow{2}{*}{Scale1} & \multirow{2}{*}{Scale2} & \multirow{2}{*}{Scale3} & \multicolumn{2}{c}{UCF-QNRF} \\
  &  &  &  & MAE & MSE \\ \hline
\multirow{3}{*}{} & \checkmark &   &   & 85.45 & 145.74 \\
 & \checkmark & \checkmark &  & 84.07 & 135.63 \\
 & \checkmark & \checkmark & \checkmark & 82.42 & 130.04 \\ 
 \hline
 \multirow{3}{*}{\checkmark} & \checkmark &  &   & 83.81 & 140.19 \\
 & \checkmark & \checkmark &  & 82.71  & 130.29 \\
 & \checkmark & \checkmark & \checkmark & \textbf{77.12}  & \textbf{124.25} \\ \hline
\end{tabular}
}
\vspace{-1mm}
\end{table}

\subsection{Ablation Studies}

\begin{figure}[t]
\centerline{
  {\footnotesize (a)} 
  \includegraphics[height=0.27\linewidth]{Figures/image_intra.png}    
  {\footnotesize (b)}
  \includegraphics[height=0.27\linewidth]{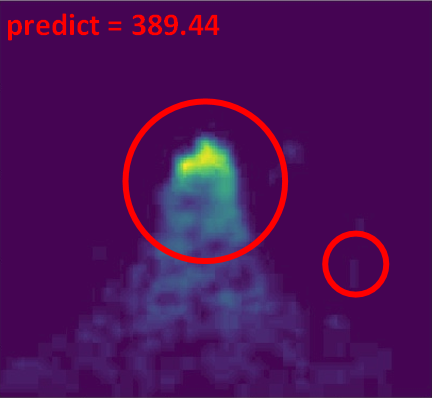}  
  {\footnotesize (c)}
  \includegraphics[height=0.27\linewidth]{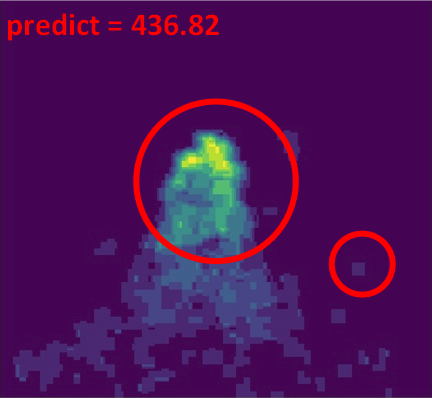}
\vspace{-2mm}
}
\caption{Visualization results of the SACC-Net (b) without and (c) with intra-fusion, where (a) shows the input image with GT = 429 from ShanghaiTech Part A; (c) is better than (b) both subjectively and objectively.}
\label{fig:intra}
\vspace{-4mm}
\end{figure}

To demonstrate the effectiveness of our fusion approach, we conducted an ablation study on how the addition of ``fusion'' and the number of scales used to improve crowd counting accuracy. Most objects in the UCF-QNRF dataset are smaller than those in other datasets. Thus, UCF-QNRF is adopted here to fairly evaluate the effect of our proposed fusion module.
In Table \ref{tab:ablation}, we can see that using our fusion module is significantly better than not using it. For example, our SACC-Net with this module reduces the error rates significantly from 85.42 to 77.12 in MAE and from 145.44 to 124.25 in MSE for the UCF-QNRF dataset.

We also evaluated the effects of the number of scales on improving the accuracy of crowd count. There are five pooling layers created in VGG19 that cause the original image to be scaled down to only 1/32$\times$1/32 ratio.  The feature map in the last layer cannot provide enough information to calculate the required covariance matrix. The first layer is too primitive for crowd counting. Since three layers provide optimal performance results,  we set $S$ to three in Eq.~\eqref{eq:all-density}. Table~\ref{tab:ablation} shows the accuracy comparisons between three combinations of three scales (corresponding to layer 2, layer 3, and layer 4).  The three-scale scale-aware loss function significantly improves the accuracy of crowd counting on the UCF-QNRF dataset, especially in the MAE metric.

{\bf Visualization results of heat maps:}  Fig.~\ref{fig:visualization_compar_loss} shows the visualization results when different loss functions were used. The ground truth of head counting in (a) is 855.  The heat maps generated by the MSE loss and the Bayesian loss~\cite{ma2019bayesian} are visualized in (b) and (c) with the predicted results of 772.69 and 901.30, respectively.  Clearly, the MSE loss performs better than the BL method. (d) is the visualization result generated by our scale-aware loss with the predicted value 884.1.  Compared to (b) and (c), our loss function can generate a more detailed heat map for heads, since annotation errors are taken into account in crowd modeling.  NoiseCC~\cite{wan2020modeling} also points out that noise from annotation will affect the accuracy of crowd counting.  
Fig.~\ref{fig:intra} shows the visualization results generated by our method wo/ and w/ IFM. The detailed heat map of smaller heads generated in (c) leads to better crowd counting accuracy, which justifies the effectiveness of IFM. The SMF can generate various synthetic layers to construct better density maps for crowd counting.
Refer to the supplementary for detailed performance evaluations of our lightweight model.

\section{Conclusions}

We presented a scale-aware SACC-Net and a new loss function that addresses the annotation noise {\em w.r.t.} scale for improving crowd counting. To deal with the scale truncation problem, The proposed SFM handles the scale truncation problem and generates a smoother scale space such that large objects can be accurately counted. The IFM is developed to fuse all feature layers within the same convolution block to generate finer-grained information for small object counting. The SACC-Net is lightweight, efficient and accurate. We also evaluated the effects of annotation variance $\alpha$ and annotation error variance $\beta$ regarding MAE. The SACC-Net outperforms other SoTA methods on four datasets. 

{\bf Future Work} includes the automatic selection of the parameters $\alpha$ and $\beta_{s}$ from data-driven learning. Also further lightweight improvement can enable the deployment of SCAA-Net to directly operate on drones.


\bibliography{aaai24}

\end{document}


\title{Scale-Aware Crowd Count Network with Annotation Error Correction - Supplementary Material}
\author{Anonymous Authors\\}
\maketitle

\begin{figure*}[t]
\centerline{
\includegraphics[width=0.8 \linewidth]{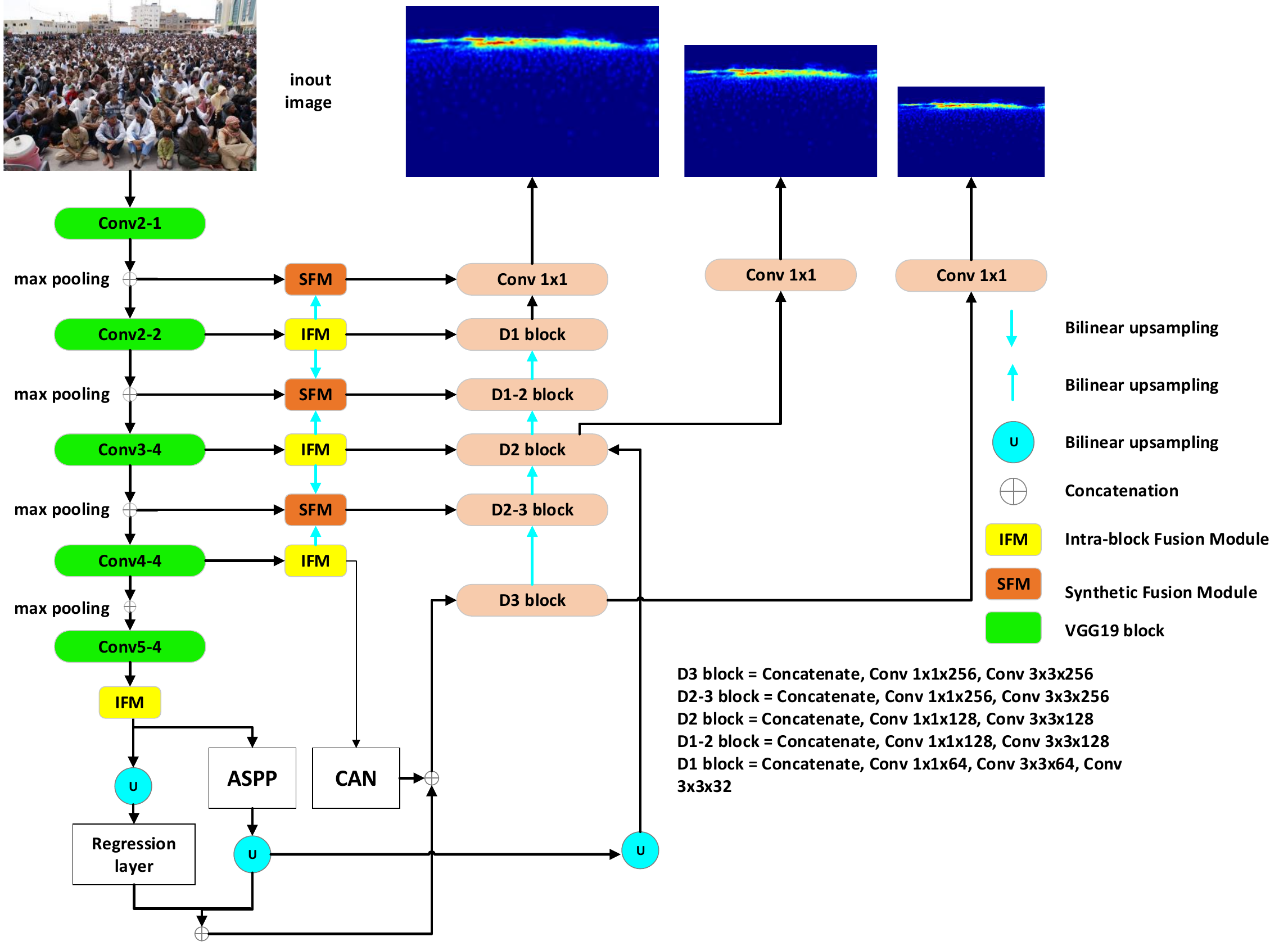}
}
\caption{Architecture of our SACC-Net for scale-aware crowd counting.}
\label{fig:Scale-IFCnet}
\end{figure*}

\section{Light-Weight Architecture}

\begin{figure*}[t]
    \centering
    \includegraphics[width = \textwidth]{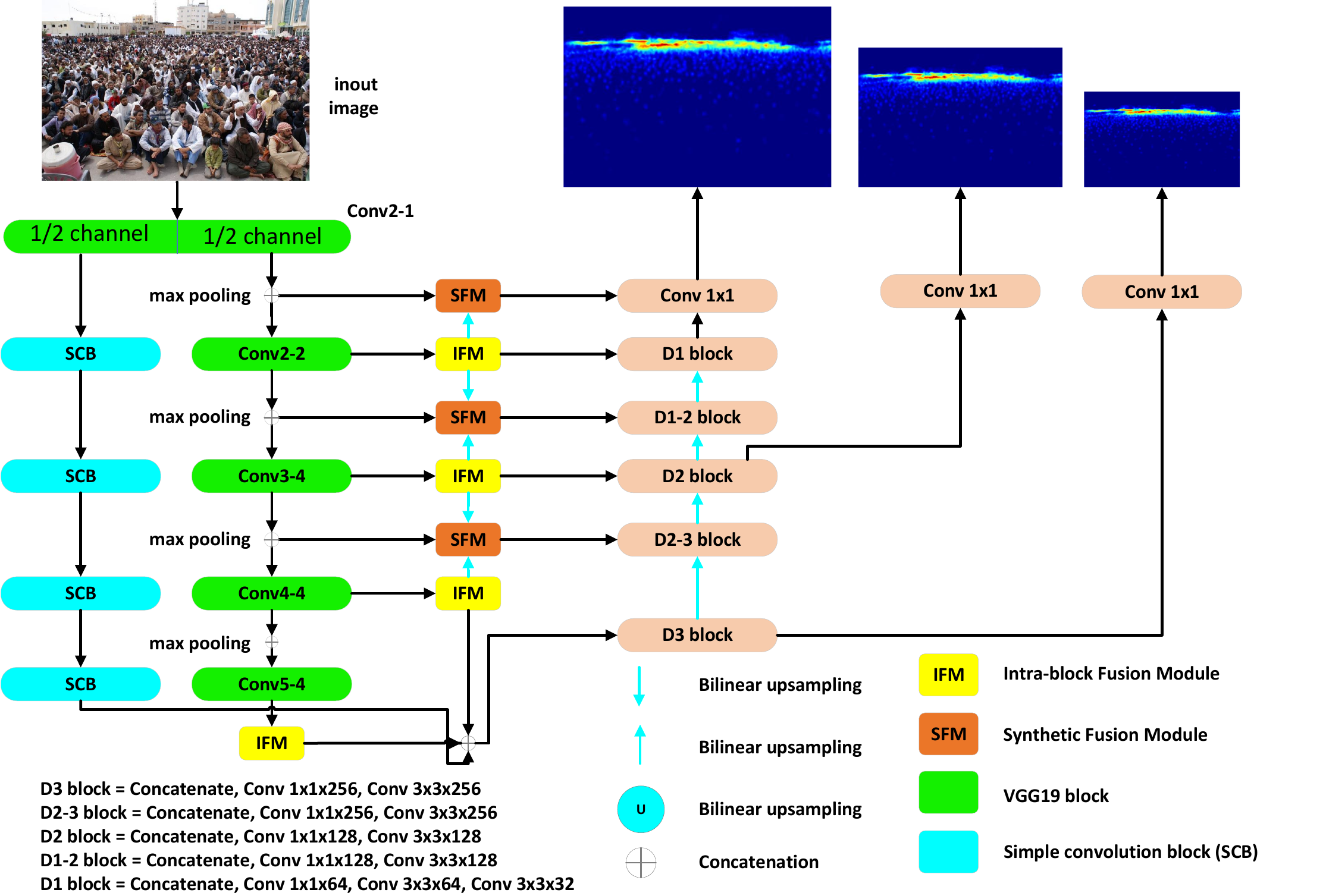}
 \vspace{0.1cm}   
    \caption{ The light-weight architecture of our SACC-Net.}
    \label{fig:LightWeightArchitecture}
\end{figure*}

\begin{figure}[t]
\centerline{
    \includegraphics[scale = 0.37]{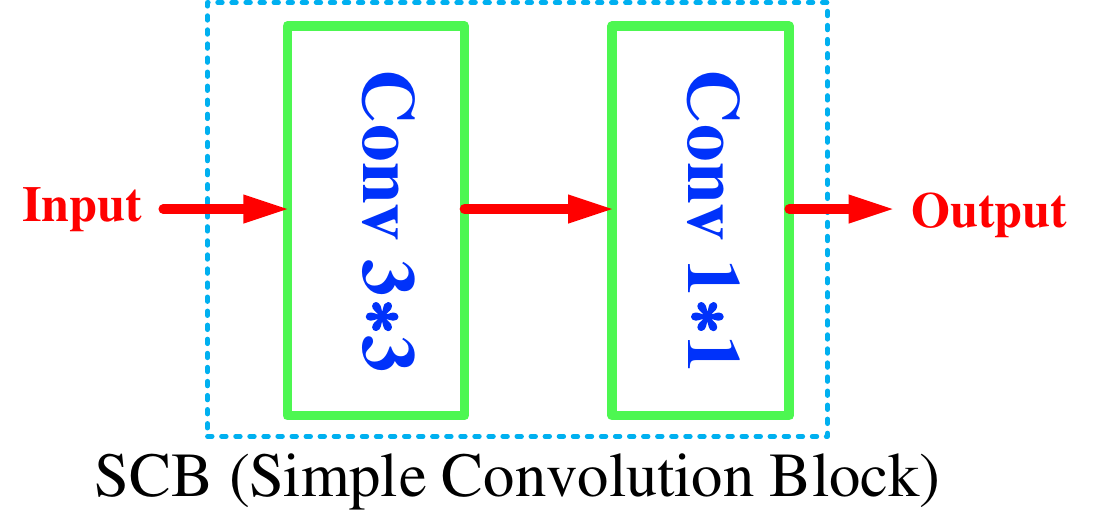}
}
\caption{ Convolutions in the Simple Convolution Block (SCB).}
\label{fig:CSP}
\end{figure}

Current density-based methods can achieve good accuracy in counting, but are inefficient for real-time applications. Figure~\ref{fig:Scale-IFCnet} shows the original design of our SACC-Net. The lightweight design here is shown in Figure~\ref{fig:LightWeightArchitecture}. Compared to the original SACC-Net, we separate the feature map into two parts: one part will pass through a VGG block and the other will pass directly through the Simple Convolution Block (SCB), which includes two simple convolutions as in Figure~\ref{fig:CSP}. The separation can not only balance the computation load of each layer, but also reduce the memory traffic load. In contrast to the original convolution on all channels, only half of the channels are sent to the next block, resulting in efficiency improvements. Thus, the model parameters can be greatly reduced while maintaining stable accuracy. 

Table~\ref{tab:COM_ALL} shows the comparative results between our light architecture and other State-of-The-Art (SoTA) methods on four benchmark datasets. We can see that our lightweight architecture performs better than other SoTA methods on four benchmark datasets. Table~\ref{tab:FPS} shows the efficiency comparisons among different backbones evaluated on a single 2080Ti GPU. Clearly, with similar accuracies, the efficiency of our light-weight version is twice that of CAN~\cite{liu2019context}, M-SFANet~\cite{thanasutives2021encoder} and SFANet~\cite{zhu2019dual}. 

\begin{table*}[t]
\caption{Performance comparisons among different SoTA crowd counting methods.
\vspace{-0.1cm}
}
\centerline{
\begin{tabular}{c|cc|cc|cc|cc}
\hline
\multirow{2}{*}{Methods}  &\multicolumn{2}{c|}{UCF-QNRF}  & \multicolumn{2}{c|}{S. H. Tech-A} & \multicolumn{2}{c|}{S. H. Tech-B} & \multicolumn{2}{c}{UCF CC 50} \\
                             & MAE           & MSE                      & MAE                & MSE                 & MAE                & MSE         & MAE                & MSE         \\ \hline
CSRNet~\cite{li2018csrnet}                             & -             & -                      & 68.2               & 115.0               & 10.3               & 16.0         & 266.1             & 397.5        \\
CAN~\cite{liu2019context}   & 107  & 183   & 62.3               & 100.0               & 7.8                & 12.2         & 212.2             & 243.7        \\
S-DCNet~\cite{xiong2019open}                            & 104.4         & 176.1            & 58.3               & 95.0                & 6.7                & 10.7        & 204.2             & 301.3         \\
SANet~\cite{cao2018scale}   & -  & -  & 67.0  & 104.5  & 8.4                & 13.6          & 258.4             & 334.9       \\
BL~\cite{ma2019bayesian}                                & 88.7          & 154.8                  & 62.8               & 101.8               & 7.7                & 12.7         & 229.3             & 308.2        \\
SFANet~\cite{zhu2019dual}  & 100.8          & 174.5                  & 59.8               & 99.3                & 6.9                & 10.9       & -             & -          \\
DM-Count~\cite{wang2020distribution}                      & 85.6          & 148.3                  & 59.7               & 95.7                & 7.4                & 11.8        & 211.0             &  291.5         \\
RPnet~\cite{zhang2015cross}                           & -          & -                 & 61.2   & 96.9   & 8.1   & 11.6       & -             & -          \\
AMSNet~\cite{Hu2020NAS}  & 101.8  & 163.2                 & 56.7.2   & 93.4   & 6.7   & 10.2     & 208.4             &  297.3            \\
M-SFANet~\cite{thanasutives2021encoder}                          & 85.6          & 151.23                   & 59.69              & 95.66              & 6.38               & 10.22         & 162.33             & 276.76       \\
TEDnet~\cite{jiang2019crowd}                             & 113.0         & 188.0                 & 64.2               & 109.1               & 8.2                & 12.8        & 249.4             & 354.5         \\
P2PNet~\cite{song2021rethinking}                              & 85.32         & 154.5                  & 52.74               & 85.06               & 6.25                & 9.9         & 172.72             & 256.18        \\
	
GauNet~\cite{cheng2022rethinking}                             & 81.6         & 153.7                  & 54.8               & 89.1               & 6.2                & 9.9          &  186.3             & 256.5       \\
MAN~\cite{Lin_2022_CVPR}                             & 77.3/83.4$^*$         & 131.5/146$^*$                   & 56.8               & 90.3               & -                & -          &  -             & -    \\  

\hline

SACC-Net (BL Loss)                           & 85.42         & 145.44                & 55.28              & 90.37               & 6.5               & 10.68         & 167.48             & 235.41       \\ 
SACC-Net (Our Loss)                              &\textbf{77.12}         & \textbf{124.25}                 & \textbf{52.19}               & \textbf{76.63}               & \textbf{6.16}                  & \textbf{9.71}  & \textbf{150.66}             & \textbf{187.89} \\
LW-SACC-Net (BL Loss)                           & 90.26         & 175.13              & 63.5              & 103.7               & 7.62               & 11.54         & 175.18             & 253.42       \\ 
LW-SACC-Net (Our Loss)                           & \textcolor{blue}{\textbf{ 81.41}}         & \textcolor{blue}{\textbf{144.51}}                & 53.74              & 88.92               & \textcolor{blue}{\textbf{6.2}}               & 10.13         & \textcolor{blue}{\textbf{157.11}}             & \textcolor{blue}{\textbf{203.64}}       
\\ 
 \hline
\end{tabular}
}
Symbol $^*$ denotes the scores re-evaluated by running the original source codes provided by the authors.
\label{tab:COM_ALL}

\end{table*}


\begin{table}[t]
\caption{Efficiency comparisons among different SoTA methods and ours using a single NVIDIA 2080Ti GPU.}
\centerline{
\setlength{\tabcolsep}{0.6mm}
\begin{tabular}{c|c|c|c}
\hline
\multirow{2}{*}{Methods} & \multicolumn{3}{c}{Frames per second (FPS)}      \\ 
\cline{2-4} 
                  & 512 × 384 & 512 × 512 & 1280 × 720 \\ \hline
CAN & 41.56     & 33.42     &  13.05  \\
M-SFANet & 42.28     & 31.45     & 12.45   \\
SFANet~\cite{zhu2019dual} & 39.71     & 30.54     & 11.16   \\ \hline
SACC-Net               & 25.24     & 20.61     & 8.19   \\
SACC-Net(light)       & \textbf{57.37}     & \textbf{45.16}     & \textbf{25.07}      \\ \hline
\end{tabular}
}
\label{tab:FPS}
\vspace{-0.4cm}
\end{table}

\begin{table*}[t]
\caption{Parameter size, MAC and FLOPS comparisons among our light-weight model, VGG16, and CAN on UCF-QNRF dataset.  The parameter size, MAC and FLOPS are computed with the input dimension 224$\times$224.}
\vspace{0.1cm}
\centering
\scalebox{1.0}
{
\begin{tabular}{c|c|c|c}
\hline
Methods    & Parameters (M)     & MAC (G)  & FLOPS (G)        \\ \hline
VGG16               & 7.89          & 15.47     & 7.73    \\
CAN         & 18.1          & 21.99  & 10.99      \\ 
M-SFANet & 28.62 &25.08  & 12.5\\
SFANet            & 17          & 19.94    & 9.9    \\ \hline
SACC-Net(Light)       & \textbf{1.86} & \textbf{6.17} & \textbf{3.0}
\\ \hline
\end{tabular}}
\label{tab:COM_MAC}
\vspace{-0.4cm}
\end{table*}

\begin{table*}[t]
\vspace{0.2cm}
\caption{Accuracy comparisons among different loss functions with various backbones on ShanghaiTech partA.}
\vspace{0.1cm}
\centering
\begin{tabular}{c|cc|cc|cc}
\hline
\multirow{2}{*}{} & \multicolumn{2}{c|}{VGG19} & \multicolumn{2}{c|}{CSRNet} & \multicolumn{2}{c}{MCNN} \\
                  & MAE         & MSE          & MAE          & MSE          & MAE         & MSE        \\ \hline
L2                & 71.4        & 136.5        & 80.61        & 149.12        & 147.82   &   201.61    \\
BL~\cite{ma2019bayesian}               & 62.8        & 101.8        &  68.2        &  115.0        & 110.2        & 173.2      \\
NoiseCC~\cite{wan2020modeling}          &  61.9        & 99.6        & 67.28         & 109.31        & 105.49       & 169.67      \\
DM-count~\cite{wang2020distribution}          &  59.7        & 95.7        & 65.71        & 105.53        & 103.81       & 165.43      \\
Gen-loss~\cite{Wan_2021_CVPR} & 61.3        &  95.4        & 63.42         & 102.51        & 102.34       & 162.87      \\ \hline
Ours              & \textbf{57.47}       & \textbf{82.29}       & \textbf{60.39}  & \textbf{96.83}       & \textbf{95.7}      & \textbf{160.12}  \\  
\hline
\end{tabular}
\label{tab:COM_diff loss with partA}
\vspace{-0.4cm}
\end{table*}

\begin{table*}[t]
\caption{Accuracy comparisons among different loss functions
with various backbones on ShanghaiTech partB.}
\vspace{0.1cm}
\centering
\begin{tabular}{c|cc|cc|cc}
\hline
\multirow{2}{*}{} & \multicolumn{2}{c|}{VGG19} & \multicolumn{2}{c|}{CSRNet} & \multicolumn{2}{c}{MCNN} \\
                  & MAE         & MSE          & MAE          & MSE          & MAE         & MSE        \\ \hline
L2                & 9.1       & 13.9        & 11.63        & 17.57        & 29.17        & 52.33      \\
BL~\cite{ma2019bayesian}               & 8.62        & 13.56        & 10.6         & 16.0
        & 26.4       &  41.3      \\
NoiseCC~\cite{wan2020modeling}          & 8.37        & 13.13        & 10.47         &15.69         &25.33        &41.04       \\
DM-count~\cite{wang2020distribution}          & 7.4        & 11.8        & 9.76        &13.82         & 23.91        &35.49       \\
Gen-loss~\cite{Wan_2021_CVPR} & 7.3         & 11.7         & 9.51         &  13.66       &  22.89     &    33.77   \\ \hline
Ours              & \textbf{7.1}       & \textbf{10.4}       & \textbf{9.38}  & \textbf{13.23}       & \textbf{20.51}      & \textbf{31.26} \\
\hline  
\end{tabular}
\label{tab:COM_diff loss with partB}
\end{table*}

\begin{table*}[t]
\caption{Ablation study of the effects of different fusion modules on our model with input dimension 512$\times$512. The SFM denotes the synthetic fusion module, and IFM denotes the intra-block fusion module.}
\centering
\setlength{\tabcolsep}{0.5mm}{
\begin{tabular}{c|l|l|ll|ll|ll|l}
\hline
\multirow{2}{*}{Methods}   & \multicolumn{1}{c|}{\multirow{2}{*}{SFM}} & \multicolumn{1}{c|}{\multirow{2}{*}{IFM}} & \multicolumn{2}{l|}{UCF-QNRF} & \multicolumn{2}{l|}{S. H. Tech-A} & \multicolumn{2}{l|}{S. H. Tech-B} & \multicolumn{1}{c}{\multirow{2}{*}{Parameters}} \\ \cline{4-9}
                           & \multicolumn{1}{c|}{}                     & \multicolumn{1}{c|}{}                     & MAE           & MSE           & MAE             & MSE             & MAE             & MSE             & \multicolumn{1}{c}{}                            \\ \hline
\multirow{4}{*}{SACC-Net} &           \checkmark                                &        \checkmark                                   &       77.12        &      124.25         &      52.19           &      76.63           &   6.16              &    9.71             &  51.2 M                                                \\ 
                           &                \ding{53}                           &          \checkmark                                 &    81.47           &      132.58         &         54.3        &       83.71          &   6.22              &         9.85        &   44.04 M                                               \\  
                           &                 \checkmark                          &      \ding{53}                                        & 82.81              &    137.62           &     54.83            &        90.39         &   6.28              &      9.93           &         32.61 M                                         \\  
                           &                      \ding{53}                        &     \ding{53}                                        &        84.16       &     149.81          &       57.5          &     98.12            &   6.35              &      10.05           &                         28.61 M                         \\ \hline
\end{tabular}}
\label{tab:ablation}
\end{table*}
\begin{table*}[t]
\caption{Ablation study of the Bayesian loss (BL) and NoiseCC loss (NoiseCC) using the Adaptive Gaussian Kernel on SACC-Net, with an input dimension of $512 \times 512$ and a training epoch of 300. }
\centering
\setlength{\tabcolsep}{0.5mm}{
\begin{tabular}{c|c|ll|ll|ll|ll}
\hline
\multirow{2}{*}{Methods}   & \multicolumn{1}{c|}{\multirow{2}{*}{Adaptive Gaussian Kernel}} & \multicolumn{2}{l|}{UCF-QNRF} & \multicolumn{2}{l|}{S. H. Tech-A} & \multicolumn{2}{l|}{S. H. Tech-B} & \multicolumn{2}{l}{UCF-CC50} \\ \cline{3-10}
                           & \multicolumn{1}{c|}{}                                          & MAE           & MSE           & MAE             & MSE             & MAE             & MSE             & MAE             & MSE                           \\ \hline
\multirow{2}{*}{BL} &           \checkmark                                &       \textcolor{red}{157.19}        &      \textcolor{red}{227.53}         &      \textcolor{red}{152.97}           &      \textcolor{red}{206.43}           &   \textcolor{red}{86.67}              &    \textcolor{red}{129.33}             &  \textcolor{red}{232.85} & \textcolor{red}{273.85}                                               \\ 
                           &                \ding{53}                           &    181.77           &      252.85         &         174.3        &       253.97          &   126.23              &         179.78        &   290.60 & 345.02                                               \\  \hline
\multirow{2}{*}{NoiseCC}                           &                 \checkmark                         & \textcolor{red}{142.87}              &    \textcolor{red}{187.14}           &     \textcolor{red}{134.83}            &        \textcolor{red}{190.11}         &   \textcolor{red}{69.24}              &      \textcolor{red}{99.91}           &         \textcolor{red}{219.64} & \textcolor{red}{255.41}                                         \\  
                           &                      \ding{53}                        &        164.71       &     229.86          &       157.52          &     238.77            &   96.31              &      150.71           &                         284.79   & 310.37                         \\ \hline
\end{tabular}}
\label{tab:adaptive ablation}
\end{table*}

\subsection{Detailed Operations in the Synthetic Fusion Module}

\begin{figure}[t]
\centerline{
    \includegraphics[scale = 0.25]{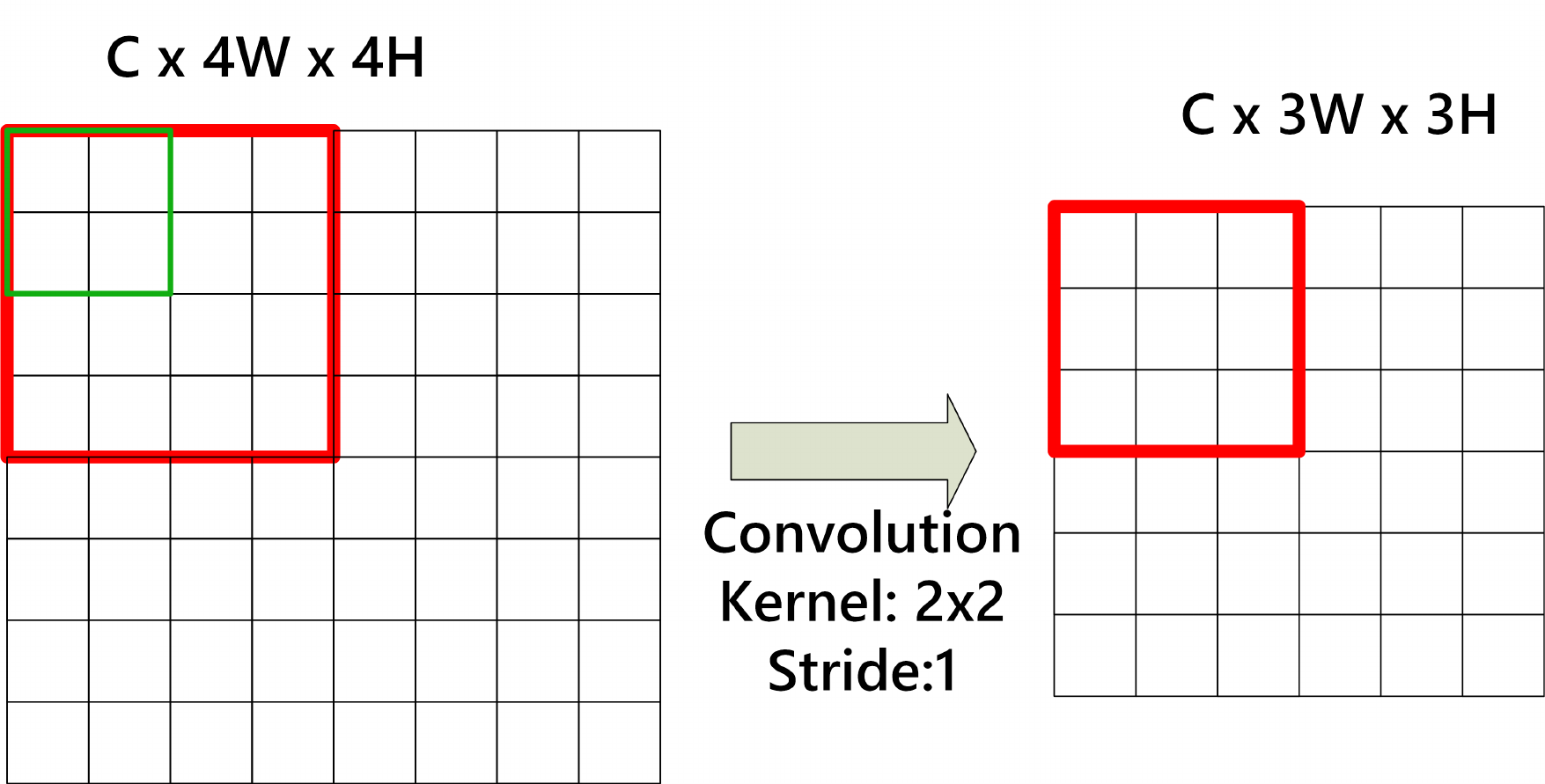}
}
\caption{ The Scaling-Down process.}
\label{fig:InterpolationDown}
\end{figure}

The pooling operation used in CNN backbone always down-samples the image dimension to half, and makes the density map scaled to $\frac{1}{2}$, $\frac{1}{4}$, $\frac{1}{8}$, ... in both the $x$ and $y$ directions. This scaling truncation is not friendly to generate heat maps with continuous scale changes. This paper introduces an inter-layer fusion to make the density map scaled to $\frac{1}{2}$, $\frac{1}{3}$, $\frac{1}{4}$, $\frac{1}{6}$, and so on. Then, a smoother scale space for fitting the ground truth whose scale changes continuously to reduce the effects of scale truncation. This module interpolates an extra layer with only $1\over 3$ scaling change as a buffer to reduce the effect of pooling operation on scale drop.  To achieve this goal, two layers (scaled by $1\over {2^i}$ and $1 \over {2^{i+1}}$) are fused together to generate another new layer (scaled by $ 1 \over {3 \times 2^{i-1}}$) by two operations, {\em i.e.}, Interpolation-Down and Interpolation-Up. 

\begin{figure}[t]
    \centering
    \includegraphics[scale = 0.26]{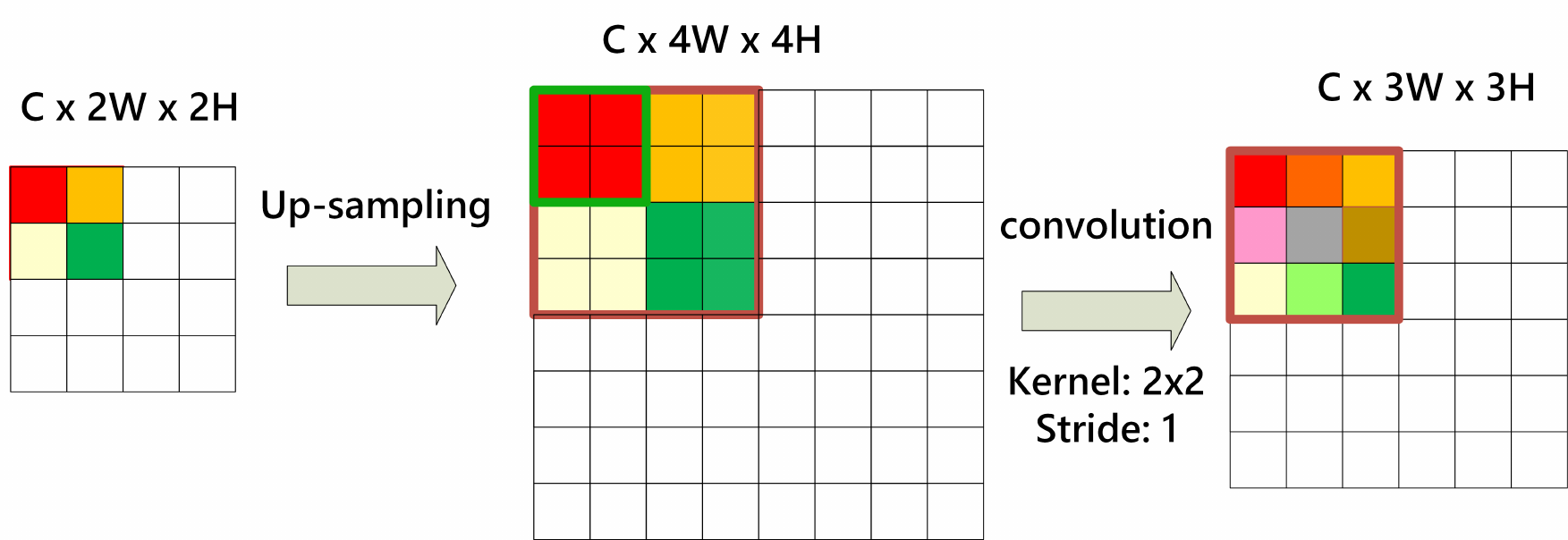}
    \caption{The Interpolation-Up process.}
    \label{fig:Interpolation-Up}
\end{figure}

 {\bf Interpolation-Down:}   As shown in Figure \ref{fig:InterpolationDown}, the input feature map of size $W \times H $ is first disassembled to regular $4 \times 4$ patches.  Then, a convolution operation with kernel $2\times 2$ and stride 1 is applied to obtain a new feature patch with the size $3 \times 3$.  This process reduces the feature map with the dimension from $C\times4W\times4H$ to $C\times 3W\times3H$.
 
{\bf Interpolation-Up:}  As shown in Figure \ref{fig:Interpolation-Up}, for each $2\times 2$ path of feature map is first up-sampled to a $4 \times 4$ patch.  Then, a convolution operation with kernel $2\times 2$ and stride 1 is applied to this $4 \times 4$ patch to obtain a new $3 \times 3$ feature map.  After that, the feature map with dimension $2W \times 2H$ is enlarged to dimension $3W \times 3H$.
\section{Time complexities of SFM and IFM}
 Both time complexities for SFM and IFM are $O (W\times H \times C)$, where $C$ is the number of channels used, and $W$ and $H$ are the input width and height.

\section{Effects of variance parameters $\alpha$ and $\beta_s$}

\begin{figure}[t]
\centerline{
  \includegraphics[width=0.9\linewidth]{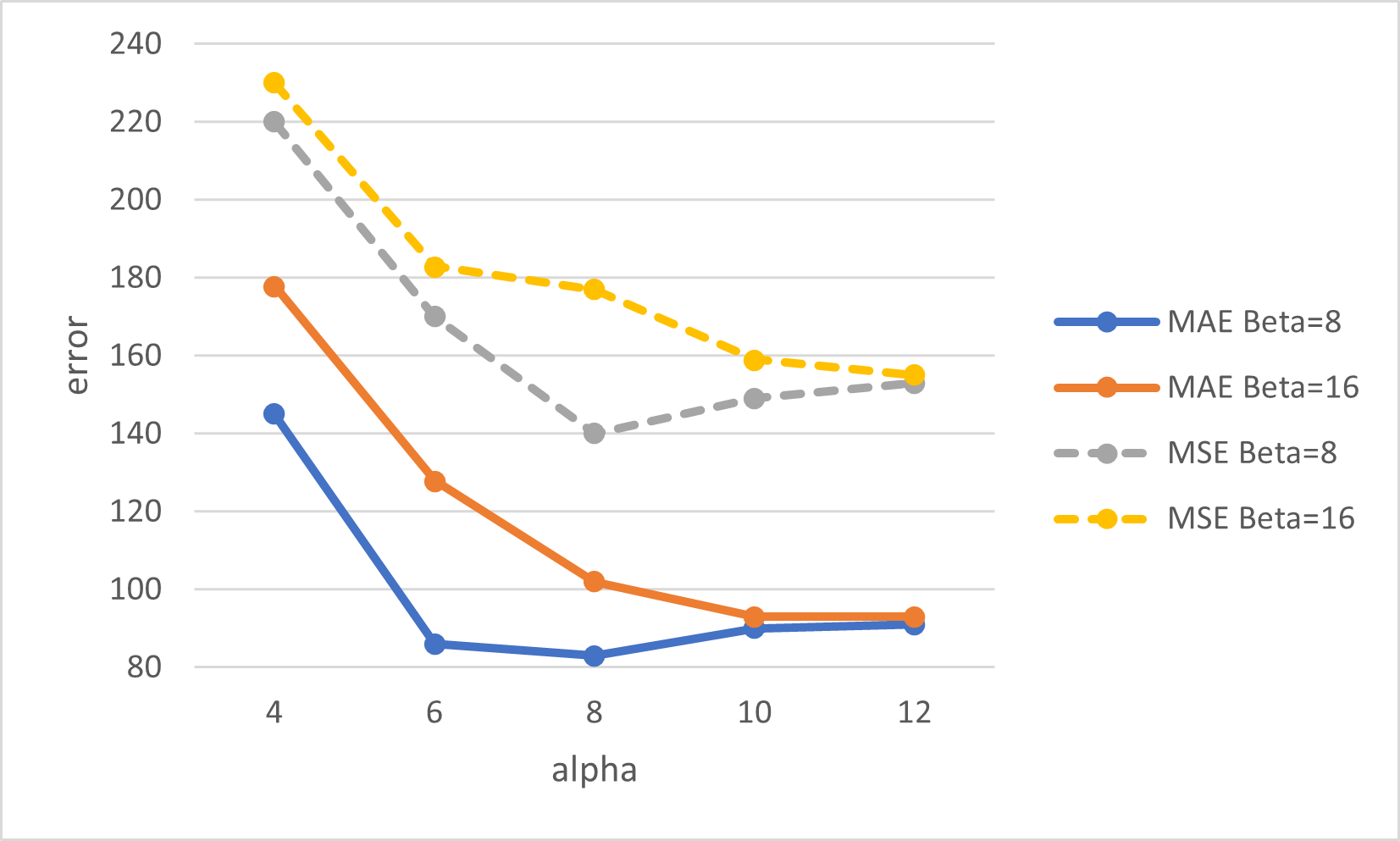}}
\caption{
 MAE changes when different annotation variance $\alpha$ and annotation error variance $\beta_{1}$ are adopted. Clearly, when $\beta_{1}=8$  and $\alpha=8$, the MAE is the lowest.
}
\label{fig:effect alpha}
\end{figure}

We also carried out an experiment to investigate the effects of annotation variance $\alpha$  and annotation error variance $\beta$ on feature map generation. As shown in Figure~\ref{fig:effect alpha}, when $\beta$ increases, the MAE decreases; but when $\beta\ge 8$, the MAE starts to increase instead. When $\alpha$ is small, the MAE is large; when $\alpha\ge8$, the MAE decreases and tends to be stable, so we set $\alpha=8$.

\section{Additional Experimental Results}
\label{sec:results}

Table~\ref{tab:COM_MAC} shows the ablation study for model parameter size, multiply-accumulate (MAC) and floating-point operations per second (FLOPS) among our light-weight architecture and other SoTA methods, respectively, where the input size is $3\times224\times224$.   For fair comparisons, VGG16 is used as a baseline.  Clearly, the parameter size of our light-weight model is one-tenth of other SoTA methods with comparable accuracies; see Table~\ref{tab:COM_ALL} for the results of CAN~\cite{liu2019context}. 

\begin{figure*}[t]
\centerline{
  \includegraphics[width=0.95\linewidth]{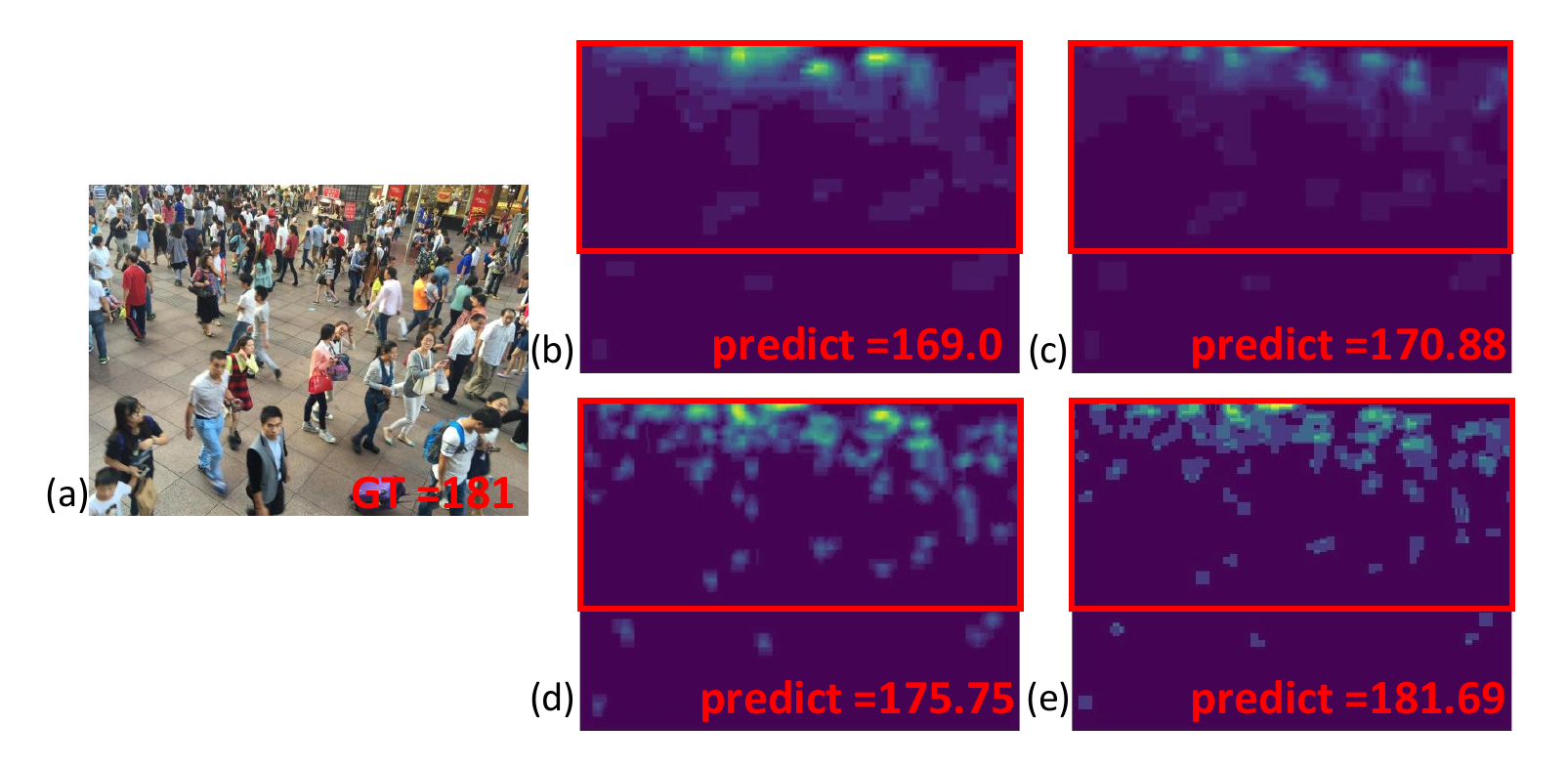}
}
\caption{Visualization results of heatmap for crowd counting generated with different loss functions. (a) The input image with groundtruth 181 heads from ShanghaiTech Part B. 
(b-d) show the crowd counting heaptmaps generated using (b) MSE loss, (c) Bayesian loss, (d) NoiseCC loss, and (e) Our scale-aware loss.
}
\label{fig:visualization_compar_loss_partB}
\end{figure*}

To verify the effectiveness of our proposed loss function,
Figure~\ref{fig:visualization_compar_loss_partB} shows the prediction maps generated by various loss functions. The ground truth number of heads here is 181 (see Figure~\ref{fig:visualization_compar_loss_partB}(a)).  The heat maps generated by the MSE loss and the Bayesian loss~\cite{ma2019bayesian} are visualized in (b) and (c) with the predicted results 169.0 and 170.88, respectively.  Clearly, the MSE loss performs better than the BL method. (d) is the visualization result of NoiseCC~\cite{wan2020modeling} with 
the predicted value 175.5.  (e) is the result generated by our scale-aware loss with the predicted value 181.69. Since our loss function considers the annotation errors in the crowd-counting model, it can generate better heat maps with better accuracy. 

\begin{figure*}[t]
\centerline{
  \includegraphics[width=0.95\linewidth]{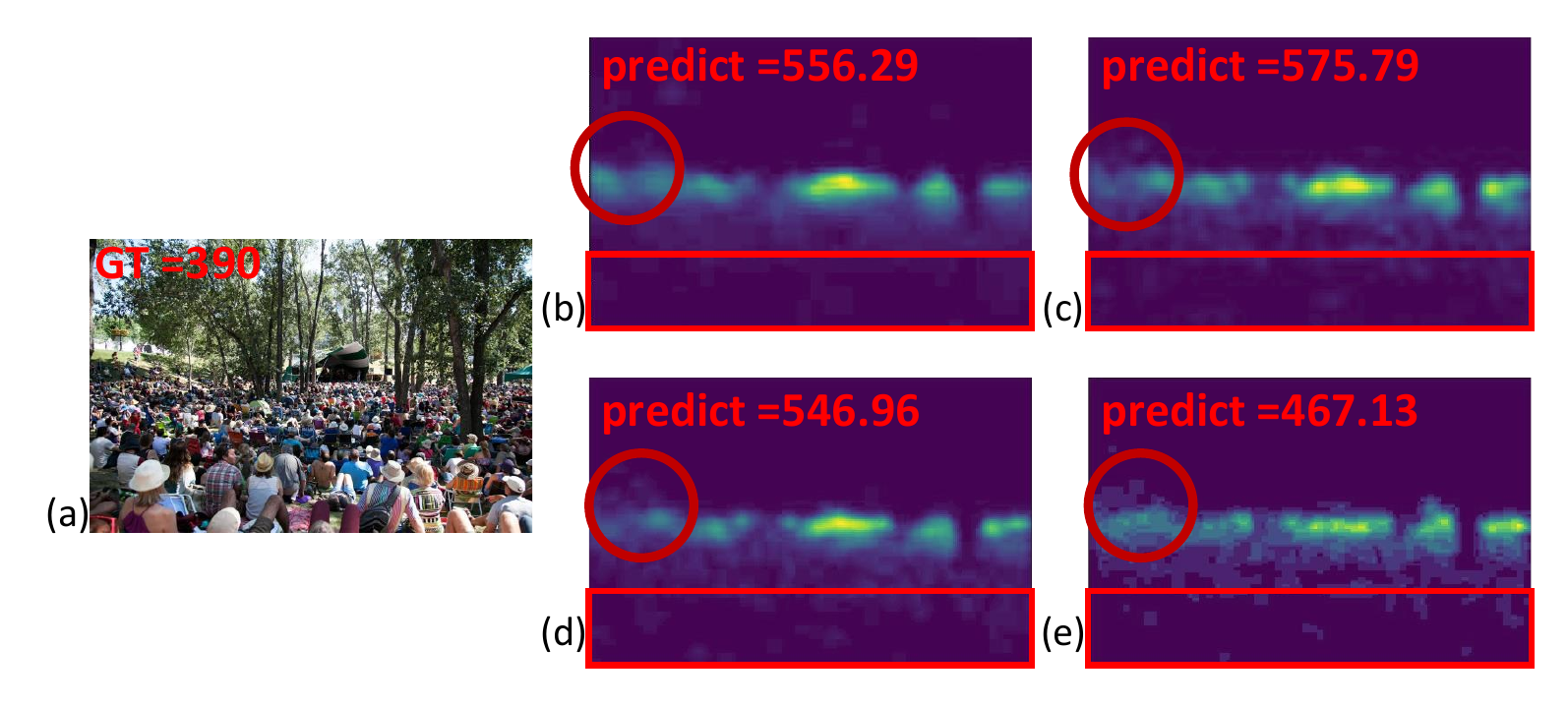}
}
\caption{Visualization results of heatmap for crowd counting generated with different loss functions. 
(a) The input image with groundtruth 390 heads from ShanghaiTech Part A. 
(b-d) show the crowd counting heaptmaps generated using (b) MSE loss, (c) Bayesian loss, (d) NoiseCC loss, and (e) Our scale-aware loss.
}
\label{fig:visualization_compar_loss}
\end{figure*}

Figure~\ref{fig:visualization_compar_loss} shows another visualization example. Clearly, our proposed method can generate a heat map with more details for counting small objects and thus yields better crowd-counting accuracy. Table~\ref{tab:ablation} shows the ablation study for the effects of SFM and IFM on our model. Clearly, IFM can improve the counting accuracy more than SFM.  However, when both fusion modules are adopted, our SACC-Net can gain the best accuracy.

\section{Scale-Adaptive Gaussian Kernel}
Table~\ref{tab:adaptive ablation} presents the ablation study of Bayesian loss (BL) and NoiseCC loss (NoiseCC) using our Scale-Adaptive Gaussian Kernel on SACC-Net, with an input dimension of $512 \times 512$ and a training epoch of 300. We can observe that both BL and NoiseCC perform better with our proposed scale-adaptive Gaussian kernel when compared to not using it, especially when the training epoch is set to 30.

\vspace{22 cm}
\bibliography{aaai24}